\RequirePackage{fix-cm}
\documentclass[twocolumn]{arxiv_svjour3}          

\smartqed  
\usepackage{graphicx}
\usepackage{ifpdf}
\usepackage{cite}
\usepackage{times}
\usepackage{epsfig}
\usepackage{graphicx}
\usepackage{amsmath}
\usepackage{amssymb}
\usepackage{booktabs}
\usepackage{amsmath}
\usepackage{multirow}
\usepackage{bm}
\usepackage{xcolor}
\usepackage{colortbl}
\usepackage{array}
\usepackage{varwidth}
\usepackage{float}
\usepackage{todonotes}
\usepackage{stfloats}
\usepackage{url}
\usepackage{xcolor}
\usepackage{changes}
\usepackage{colortbl}
\usepackage{amsmath,graphicx}
\usepackage{amsmath}
\usepackage{algorithmic}
\usepackage{array}
\usepackage{hyperref}
\hypersetup{linkcolor=blue; citecolor=green;}

\newcommand*{\rom}[1]{\expandafter\@slowromancap\romannumeral #1@}
\long\def\IGNORE#1{} \long\def\COMMENT#1{}

\usepackage{pifont}
\newcommand{\xmark}{\ding{55}}%

\usepackage{afterpage}

\begin{document}

\title{Recognizing American Sign Language Manual Signs from RGB-D Videos}


\author{Longlong~Jing$^{*}$          \and
        Elahe~Vahdani$^{*}$   \and 
        Matt~Huenerfauth \and 
        Yingli~Tian$^{\dag}$
}


\institute{L. Jing and E. Vahdani \at
              Department of Computer Science, The Graduate Center, The City University of New York, NY, 10016. \\
              \email{\{ljing,~evahdani\}@gradcenter.cuny.edu}\\
              $^*$Equal Contribution
           \and 
            M. Huenerfauth \at
            Golisano College of Computing and Information Sciences, the Rochester Institute of Technology (RIT), Rochester, NY, USA.\\ \email{matt.huenerfauth@rit.edu}\\
           \and
           Y. Tian \at
              Department of Electrical Engineering, The City College, and the Department of Computer Science, the Graduate Center, the City University of New York, NY, 10031. \\
            \email{ytian@ccny.cuny.edu}\\
            $^{\dag}$ Corresponding Author 
}

\date{}


\maketitle

\begin{abstract}
In this paper, we propose a 3D Convolutional Neural Network (3DCNN) based multi-stream framework to recognize American Sign Language (ASL) manual signs (consisting of movements of the hands, as well as non-manual face movements in some cases) in real-time from RGB-D videos, by fusing multimodality features including hand gestures, facial expressions, and body poses from multi-channels (RGB, depth, motion, and skeleton joints). To learn the overall temporal dynamics in a video,  a proxy video is generated by selecting a subset of frames for each video which are then used to train the proposed 3DCNN model. We collect a new ASL dataset, ASL-100-RGBD, which contains $\mathit{42}$ RGB-D videos captured by a Microsoft
Kinect V2 camera, each of $\mathit{100}$ ASL manual signs, including RGB channel, depth maps, skeleton joints, face features, and HDface. The dataset is fully annotated for each semantic region (i.e. the time duration of each word that the human signer performs). Our proposed method achieves $\mathit{92.88\%}$ accuracy for recognizing $\mathit{100}$ ASL words in our newly collected ASL-100-RGBD dataset. The effectiveness of our framework for recognizing hand gestures from RGB-D videos is further demonstrated  on the Chalearn IsoGD dataset and achieves $76\%$ accuracy which is $5.51$\% higher than the state-of-the-art work in terms of average fusion by using only 5 channels instead of 12 channels in the previous work. 

\end{abstract}

\section{Introduction}
The focus of our research is to develop a real-time system that can automatically identify ASL manual signs (individual words, which consist of movements of the hands, as well as facial expression changes) from RGB-D videos.  However, our broader goal is to create useful technologies that would support ASL education, which would utilize this technology for identifying ASL signs and provide ASL students immediate feedback about whether their signing is fluent or not.  

There are more than one hundred sign languages used around the world, and ASL is used throughout the U.S. and Canada, as well as other regions of the world, including West Africa and Southeast Asia. Within the U.S.A., about $28$ million people today are Deaf or Hard-of-Hearing (DHH) \cite{Ncra}. There are approximately $500,000$ people who use ASL as a primary language \cite{mitchell2006many}, and since there are significant linguistic differences between English and ASL, it is possible to be fluent in one language but not in the other. 

In addition to the many members of the Deaf community who may prefer to communicate in ASL, there are many individuals who seek to learn the language. Due to a variety of educational factors and childhood language exposure, researchers have measured lower levels of English literacy among many deaf adults in the U.S. \cite{traxler2000stanford}. Studies have shown that deaf children raised in homes with exposure to ASL have better literacy as adults, but it can be challenging for parents, teachers, and other adults in the life of a deaf child to rapidly gain fluency in ASL. The study of ASL as a foreign language in universities has significantly increased by $16.4$\% from $2006$ to $2009$, which ranked ASL as the 4th most studied language at colleges \cite{Furman10}. Thus, there are many individuals would benefit from a flexible way to practice their ASL signing skills, and our research investigates technologies for recognizing signs performed in color and depth videos, as discussed in \cite{huenerfauth2017evaluation}.  

While the development of user-interfaces for educational software was described in our prior work \cite{huenerfauth2017evaluation}, this article instead focuses on the development and evaluation of our ASL recognition technologies, which underlie our educational tool.  Beyond this specific application, technology to automatically recognize ASL signs from videos could enable new communication and accessibility technologies for people who are DHH, which may allow these users to input information into computing systems by performing sign language or may serve as a foundation for future research on machine translation technologies for sign languages. 

The rest of this article is structured as follows:  Section 1.1 provides a summary of relevant ASL linguistic details, and Section 1.2 motivates and defines the scope of our contributions. Section 2 surveys related work in ASL recognition, gesture recognition in videos, and some video-based ASL corpora (collections of linguistically labeled video recordings). Section 3 describes our framework for ASL recognition, Section 4 describes the new dataset of 100 ASL words captured by a RGBD camera which is used in this work, and Section 5 presents the experiments to evaluate our ASL recognition model and the extension of our framework for Chalearn IsoGD dataset. Finally, Section 6 summarizes the proposed work.

\subsection{ASL Linguistics Background}

ASL is a natural language conveyed through movements and poses of the hands, body, head, eyes, and face \cite{Valli11}. Most ASL signs consist of the hands moving, pausing, and changing orientation in space.  Individual ASL signs (words) consist of a sequence of several phonological segments, which include:

\vspace{-2 mm}
\begin{itemize}
\itemsep0em

\item An essential parameter of a sign is the configuration of the hand, i.e. the degree to which each of the finger joints are bent, which is commonly referred to as the ``handshape."  In ASL, there are approximately $86$ handshapes which are commonly used \cite{Neidle12}, and the hand may transit between handshapes during the production of a single sign.

\item During an ASL sign, the signer's hands will occupy specific locations and will perform movement through space. Some signs are performed by a single hand, but most are performed using both of the signer's hands, which move through the space in front of their head and torso. During two-handed signs, the two hands may have symmetrical movements, or the signer's dominant hand (e.g. the right hand of a right-handed person) will have greater movements than the non-dominant hand. 

\item The orientation of the palm of the hand in 3D space is also a meaningful aspect of an ASL sign, and this parameter may differentiate pairs of otherwise identical signs.  

\item Some signs co-occur with specific ``non-manual signals," which are generally facial expressions that are characterized by specific eyebrow movement, head tilt/turn, or head movement (e.g., forward-backward relative to the torso).  

\end{itemize}

\begin{figure}[t]
\begin{center}
\includegraphics[width=0.48\textwidth]{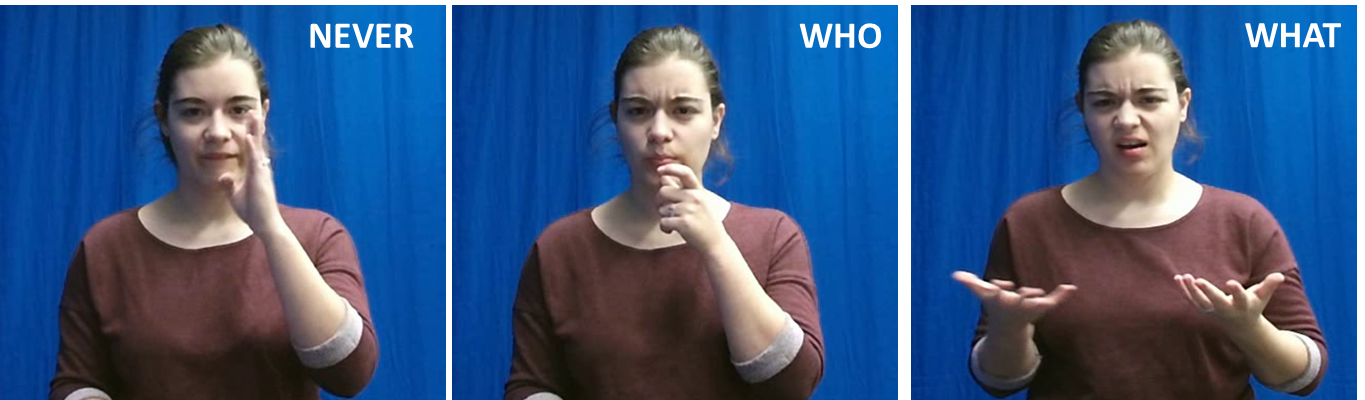}
\caption{Example images of lexical facial expressions along with hand gestures for signs: NEVER, WHO, and WHAT.  For NEVER, the signer shakes her head side-to-side slightly, which is a Negative facial expression in ASL.  For WHO and WHAT, the signer is furrowing the brows and slightly tilting moving the head forward, which is a WH-word Question facial expression in ASL.}
\label{fig:gesture+expression}
\end{center}
\end{figure}

As discussed in \cite{Mulrooney10}, facial expressions in ASL are most commonly utilized to convey information about entire sentences or phrases, and these classes of facial expressions are commonly referred to as ``syntactic facial expressions."  While some researchers, e.g. \cite{Metaxas12}, have investigated the identification of facial expressions that extend across multiple words to indicate grammatical information, in this paper, we describe our work on recognizing manual signs which  consist  of  movements of  the  hands and facial  expression changes.  

In addition to ``syntactic" facial expressions that extend across multiple words in an ASL sentence, there exists another category of facial expressions, which is specifically relevant to the task of recognizing individual signs: ``lexical facial expressions," which are considered as a part of the production of an individual ASL word (see examples in Fig. \ref{fig:gesture+expression}).  Such facial expressions are therefore essential for the task of sign recognition. For instance, words with negative semantic polarity, e.g. NONE or NEVER, tend to occur with a negative facial expression consisting of a slight head shake and nose wrinkle. In addition, there are specific ASL signs that almost always occur in a context in which a specific ASL syntactic facial expression occurs. For instance, some question words, e.g. WHO or WHAT, tend to co-occur with a syntactic facial expression (brows furrowed, head tilted forward), which indicates that an entire sentence is a WH-word Question. Thus, the occurrence of such a facial expression may be useful evidence to consider when building a sign-recognition system for such words.

\subsection{Motivations and Scope of Contributions}

As discussed in Section 2.1, most prior ASL recognition research typically focuses on isolated hand gestures of a restricted vocabulary. In this paper, we propose a 3D multi-stream framework to recognize a set of grammatical ASL words in real-time from RGB-D videos, by fusing multimodality features including hand gestures, facial expressions, and body poses from multi-channels (RGB, depth, motion, and skeleton joints). In an extension to our previous work  \cite{ChenyangZhang16} and \cite{Yuancheng18}, the main contributions of the proposed framework can be summarized as follows:
\vspace{-2 mm}
\begin{itemize}
\itemsep0em

\item We propose a 3D multi-stream framework by using 3D convolutional neural network for ASL recognition in RGB-D videos by fusing multi-channels including RGB, depth, motion, and skeleton joints. 

\item We propose a random temporal augmentation strategy to augment the training data to handle wide diverse videos of relative small datasets. 

\item We create a new ASL dataset, ASL-100-RGBD, including multiple modalities (facial movements, hand gestures, and body pose) and multiple channels (RGB, depth, skeleton joints, and HDface) by collaborating with ASL linguistic researchers; this dataset contains annotation of the time duration when each ASL word is performed by the human in the video. The dataset will be released to public with the publication of this article.

\item We further evaluate the proposed framework to recognize hand gestures on the Chalearn LAP IsoGD dataset \cite{LargeScaleRGBDHandGesture16} which consists of $249$ gesture classes in RGB-D videos. The accuracy of our framework is $5.51$\% higher than the state-of-the-art work in terms of average fusion using fewer channels ($5$ channels instead of $12$). 

\end{itemize}

\section{Related Work}

\subsection{RGB-D Based ASL Recognition}

Sign language (SL) recognition has been studied for three decades since the first attempt to recognize Japanese SL by Tamura and Kawasaki in 1988 \cite{Tamura88}. The existing SL recognition research can be classified as sensor-based methods including data gloves and body trackers to capture and track hand and body motions \cite{Kadous96,Liang98, Fang07, Kong14} and non-intrusive camera-based methods by applying computer vision technologies \cite{Starner98,YangSpotting09, YangSarkar10, DanielKelly10, pigou2018beyond,  Pigou14, huang2018video, pu2018dilated, camgoz2018neural, pigou2017gesture, cui2017recurrent, camgoz2017subunets, koller2015deep, koller2016deep, liu2016real, gattupalli2016evaluation, koller2018deep, charles2014automatic }.  While much research in this area focuses on the hands, there is also some research focusing on linguistic information conveyed by the face and head of a human performing sign language, such as  \cite{JingjingLiu13, Metaxas12, kumar2018independent, vonAgris08}. More details about SL recognition can be found in these survey papers \cite{Ong05, Er-Rady17}. 

As cost-effective consumer depth cameras have become available in recent years, such as RGB-D cameras of Microsoft Kinect V2 \cite{MSKinect2}, Intel Realsense \cite{IntelRealSense}, Orbbec Astra \cite{OrbbecAstra}, it has become practical to capture high resolution RGB videos and depth maps as well as to track a set of skeleton joints in real time. Compared to traditional 2D RGB images, RGB-D images provide both photometric and geometric information. Therefore, recent research work has been motivated to investigate ASL recognition using both RGB and depth information \cite{Pugeault11, Zafrulla11, Chai13, RenYuan13, ChenyangZhang16, Yuancheng18,  Jiang14, Almeidaab14, YangSensors15, buehler2011upper}. In this article, we briefly summarize ASL recognition methods based on RGB-D images or videos.

Some early work of SL recognition based on RGB-D cameras only focused on a very small number of signs from static images \cite{Pugeault11, RenYuan13, Keskin12}. Pugeault and Richard proposed a multi-class random forest classification method to recognize $24$ static ASL fingerspelling alphabet letters by ignoring the letters $\mathit{j}$ and $\mathit{z}$ (as they involve motion) and by combining both appearance and depth information of handshapes captured by a Kinect camera \cite{Pugeault11}. Keskin \textit{et al.} \cite{Keskin12} recognized $24$ static handshapes of the ASL alphabet, based on scale invariant features extracted from depth images, and then fed to a Randomized Decision Forest for classification at the pixel level, where the final recognition label was voted based on a majority. Ren et al. proposed a modified Finger-Earth Mover’s Distance metric to recognize static handshapes for $10$ digits captured using a Kinect camera \cite{RenYuan13}. 

While these systems only used the static RGB and depth images, some studies employed the RGB-D videos for ASL recognition. Zafrulla \textit{et al.} developed a hidden Markov model (HMM) to recognize $19$ ASL signs collected by a Kinect and compared the performance with that from colored-glove and accelerometer sensors \cite{Zafrulla11}. For the Kinect data, they also compared the system performance between the signer seated and standing and found that higher accuracy resulted when the users were standing. Yang developed a hierarchical conditional random field method to recognize $24$ manual ASL signs (seven one-handed and $17$ two-handed) from the handshape and motion in RGB-D videos \cite{YangSensors15}. Lang \textit{et al.} \cite{Lang12} presented a HMM framework to recognize $25$ signs of German Sign Language using depth-camera specific features. Mehrotra \textit{et al.} \cite{Mehrotra15} employed a support vector machine (SVM) classifier to recognize $37$ signs of Indian Sign Language based on 3D skeleton points captured using a Kinect camera. Almeida et al. \cite{Almeidaab14} also employed a SVM classifier to recognize $34$ signs of Brazilian Sign Language using handshape, movement and position captured by a Kinect. Jiang \textit{et al.} proposed to recognize $34$ signs of Chinese Sign Language based on the color images and the skeleton joints captured by a Kinect camera \cite{Jiang14}.   Recently, Kumar \textit{et al.} \cite{Kumar17} combined a Kinect camera with a Leap Motion sensor to recognize $50$ signs of India Sign Language.    

As discussed above, SL consists of hand gestures, facial expressions, and body poses. However, most existing work has focused only on hand gestures without combining with facial expressions and body poses. While a few attempted to combine hand and face \cite{vonAgris08, YangSpotting09, Metaxas12, koller2015continuous, pigou2017gesture, kumar2018independent}, they only use RGB videos. To the best of our knowledge, we believe that this is the first work that combines multi-channel RGB-D videos (RGB and depth) with fusion of multi-modality features (hand, face, and body) for ASL recognition.

\subsection{CNN for Action and Hand Gesture Recognition}

Since the work of AlexNet \cite{AlexNet} which makes use of the powerful computation ability of GPUs, deep neural networks (DNNs) have enjoyed a renaissance in various areas of computer vision, such as image classification \cite{donahue2013decaf, szegedy2014going}, object detection \cite{girshick2014rich, he2014spatial}, image description \cite{donahue2014long, karpathy2014deep}, and others. Many efforts have been made to extend CNNs from the image to the video domain \cite{fernando2017rank}, which is more challenging since video data are much larger than images; therefore, handling video data in the limited GPU memory is not tractable. An intuitive way to extend image-based CNN structures to the video domain is to perform the fine-tuning and classification process on each frame independently, and then conduct a later fusion, such as average scoring, to predict the action class of the video \cite{KarpathyCVPR14}. To incorporate temporal information in the video, \cite{simonyan2014two} introduced a two-stream framework. One stream was based on RGB images, and the other, on stacked optical flows. Although that work proposed an innovative way to learn temporal information using a CNN structure, in essence, it was still image-based, since the third dimension of stacked optical flows collapsed immediately after the first convolutional layer. 

To model the sequential information of extracted features from different segments of a video, \cite{donahue2014long} and \cite{ng2015beyond} proposed to input features into Recurrent Neural Network (RNN)  structures, and they achieved good results for action recognition. The former emphasized pooling strategies and how to fuse different features, while the latter focused on how to train an end-to-end DNN structure that integrates CNNs with RNNs. These networks mainly use CNN to extract spatial features, then RNN is applied to extract the temporal information of the spatial features. 3DCNN was recently proposed to learn the spatio-temporal features with 3D convolution operations \cite{C3D, T3D, 3DResNet, 3D, P3D, 3DMultiModel, VideoYOLO}, and has been widely used in video analysis tasks such as video caption and action detection. 3DCNN is usually trained with fixed-length clips (usually 16 frames \cite{3DResNet},\cite{C3D},) and later fusion is performed to obtain the final category of the entire video. Hara \textit{et al.} \cite{3DResNet} proposed the 3D-ResNet by replacing all the 2D kernels in 2D-ResNet with 3D convolution operations. With its advantage of avoiding gradient vanishing and explosion, the 3D-ResNet outperforms many complex networks. 

ASL recognition shares properties with video action recognition, therefore, many networks for video action have been applied to this task. Pigou \textit{et al.} proposed temporal residual networks for gesture and sign language recognition \cite{pigou2017gesture} and  temporal convolutions on top of the features extracted by 2DCNN for gesture recognition \cite{pigou2018beyond}.  Huang \textit{et al.} proposed a Hierarchical Attention Network with Latent Space (LS-HAN) which eliminates the pre-processing of the temporal segmentation \cite{huang2018video}. Pu \textit{et al.} proposed to employ 3D residual convolutional network (3D-ResNet) to extract visual features which are then fed to a stacked dilated convolution network with connectionist temporal classification to map the visual features into text sentence \cite{pu2018dilated}. Camgoz \textit{et al.} attempted to generate spoken language translations from sign language video \cite{camgoz2018neural}. Camgoz \textit{et al.} proposed SubUNets for simultaneous hand shape and continuous sign language recognition \cite{camgoz2017subunets}.  Cui \textit{et al.} proposed a weakly supervised framework to train the network from videos with ordered gloss labels but no exact temporal locations for continuous sign language recognition \cite{cui2017recurrent}. In prior work, our research team proposed a 3D-FCRNN for ASL recognition by combining the 3DCNN and a fully connected RNN \cite{Yuancheng18}.

\begin{figure*}[t]
\begin{center}
\includegraphics[width=\textwidth]{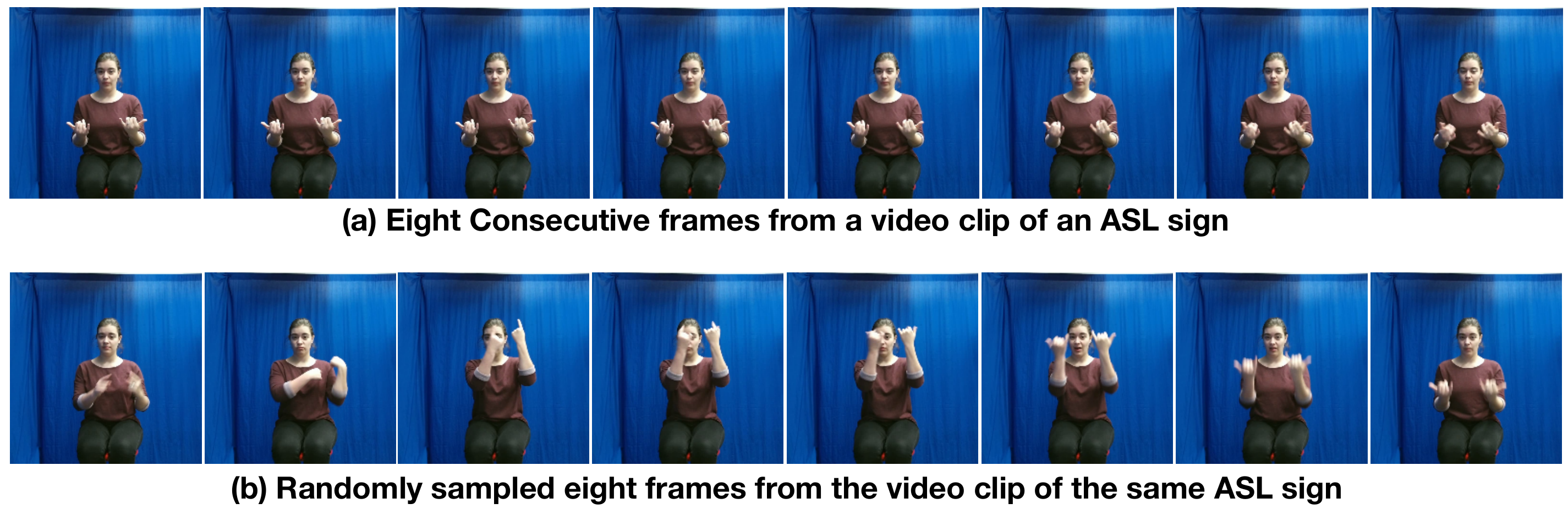}
\caption{Generating representative proxy video by our proposed random temporal augmentation.  (a) Eight consecutive frames from a video clip of an ASL sign. (b) Randomly sampled eight frames from the video clip of the same ASL sign. With the same number of frames, the proxy video reserves more temporal dynamics of the ASL sign.}
\label{fig:data_aug}
\end{center}
\end{figure*}


\subsection{Public Camera-based ASL Datasets}

As discussed in Section 2.1, technology to recognize ASL signs from videos could enable new educational tools or assistive technologies for people who are DHH, and there has been significant prior research on sign language recognition.  However, a limiting factor for much of this research has been the scarcity of video recordings of sign language that have been annotated with time interval labels of the words that the human has performed in the video: For ASL, there have been some annotated video-based datasets \cite{neidle2012new} or collections of motion capture recordings of humans wearing special sensors \cite{lu2012cuny}. Most publicly available datasets, e.g. \cite{forster2012rwth, koller2015continuous}, contain general ASL vocabularies from RGB videos and a few with RGB-D channels.

\textbf{2D Camera-based ASL databases:} The  American Sign Language Linguistic Research Project (ASLLRP) dataset contains video clips of signing from the front and side and includes a close-up view of the face \cite{neidle2012new}, with annotations for 19 short narratives (1,002
utterances) and 885 additional elicited utterances from four Deaf native ASL signers; annotation includes: the start and endpoints of each sign, a unique gloss label for each sign, part of speech, and start and end points of a range of non-manual behaviors (e.g., raised/lowered eyebrows, head position and periodic head movements, expressions of the nose and mouth) also labeled with respect to the linguistic information that they convey (serving to mark, e.g., different sentence types, topics, negation, etc.). Dreuw \textit{et al.} \cite{DreuwECCV2010} produced several subsets from the ASLLRP dataset as benchmark databases for automatic recognition of isolated and continuous sign language.

The American Sign Language Lexicon Video Dataset (ASLLVD) \cite{Athitsos08} is a large dataset of videos of isolated signs from ASL. It contains video sequences of about 3,000 distinct signs, each produced by 1 to 6 native ASL signers recorded by four cameras under three views: front, side, and face region, along with annotations of those sequences, including start/end frames and class label (i.e., gloss-based identification) of every sign, as well as hand and face locations at every frame.

The RVL-SLLL ASL Database \cite{RVL-SLLLASLDatabase02} consists three sets of ASL videos with distinct motion patterns,  distinct handshapes, and structured sentences respectively. These videos were captured from 14 native ASL signers (184 videos per signer) under different lighting conditions. For annotation, the videos  with distinct motion patterns or  distinct handshapes are saved as separate clips. However, there is no detailed annotations for the videos of structured sentences which limited the usefulness of the database. 

\textbf{RGB-D Camera-based ASL and Gesture Databases:} Recently, some RGB-D databases have been collected for hand gesture and SL recognition, for ASL or other sign languages \cite{Chai13, Pigou14, forster2012rwth}. Here we only briefly summarize RGB-D databases for ASL.  

The "Spelling-It-Out" dataset consists of 24 static handshapes of the ASL fingerspelling alphabet, ignoring the letters “j” and “z” as they involve motion, from four signers; each signer repeats 500 samples for each letter in front of a Kinect camera \cite{Pugeault11}. The NTU dataset consists of 10 static hand gestures for digits 1 to 10 and was collected from 10 subjects by a Kinect camera. Each subject performs 10 different poses with variations in hand orientation, scale, articulation for the same gesture, and there is a color image and the corresponding depth map for each \cite{RenYuan13}. 

The Chalearn LAP IsoGD dataset \cite{LargeScaleRGBDHandGesture16} is a large-scale hand gesture RGB-D dataset, which is derived from Chalearn Gesture dataset (CGD 2011) \cite{GuyonCGD2011}.  This dataset consists of $47,933$ RGB-D video clips fallen into $249$ classes of hand gestures including mudras (Hindu/ Buddhist hand gestures), Chinese numbers, and diving signals. Although it is not about ASL recognition, it can be used to learn RGB-D features from different environment settings. Using the learned features as a pretrained model, the fine-tuned ASL recognition model will be more robust to handle different backgrounds and scales (e.g. distance variations between Kinect camera and the signer).

To support our research, we have collected and annotated a novel RGB-D ASL dataset, ASL-100-RGBD, described in Section 4, with the following properties:

\vspace{-2 mm}
\begin{itemize}
\itemsep0em
\item $\mathit{100}$ ASL signs have been collected which are performed by $\mathit{15}$ individual signers (often with multiple recordings from each signer).

\item The ASL-100-RGBD dataset has been captured using a Kinect V2 camera and contains multiple channels including RGB, depth, skeleton joints, and HDface. 

\item Each video consists of the 100 ASL words with time-coded annotations in collaboration with ASL computational linguistic researchers.

\item The $\mathit{100}$ ASL words have been strategically selected to support sign recognition technology for ASL education tools (many of these words consist of hand gestures and facial expression changes), with the detailed vocabulary composition described in Section 4.

\end{itemize}


\begin{figure*}[t]
\begin{center}
\includegraphics[width=\textwidth]{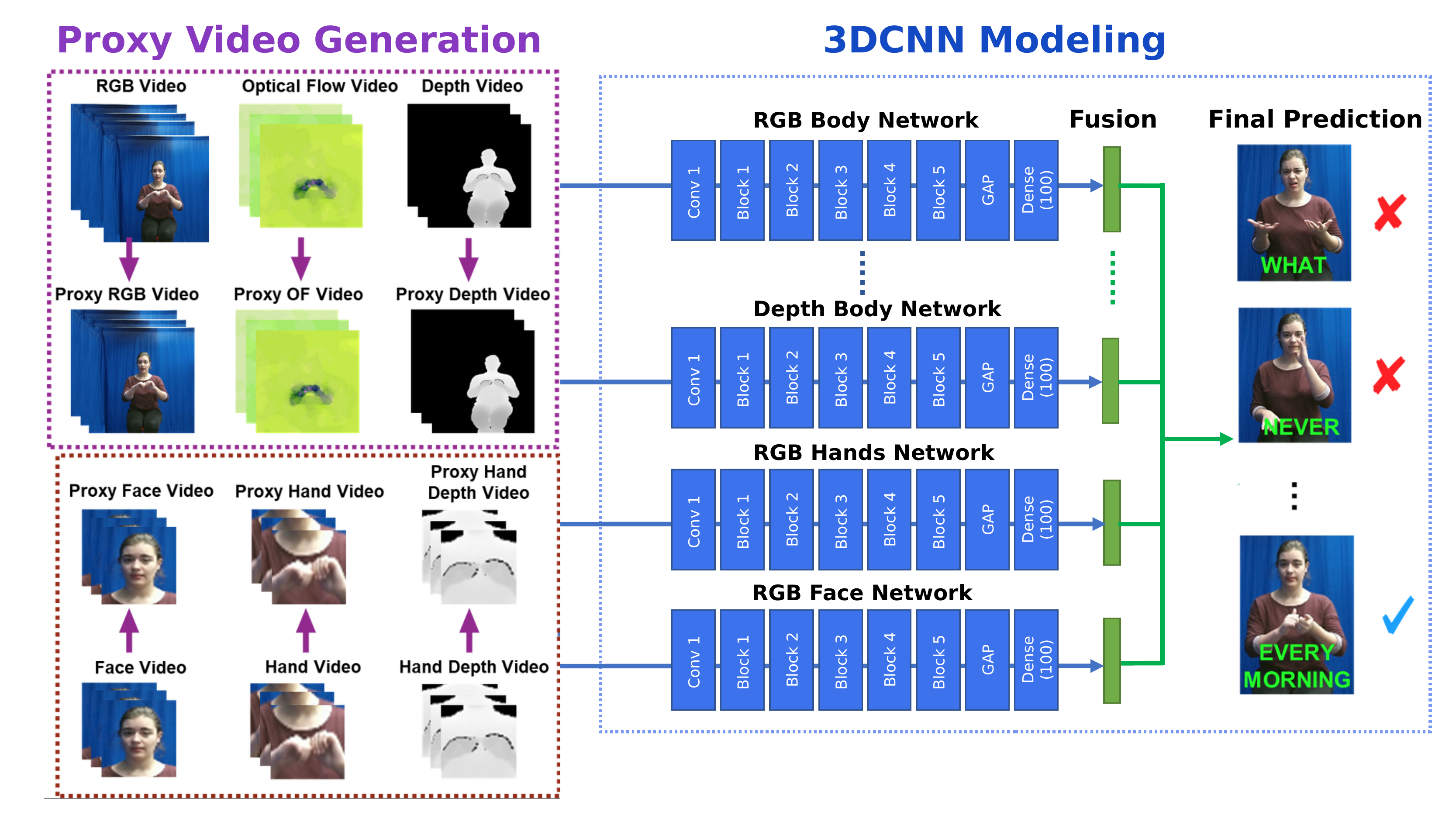}
\caption{The pipeline of the proposed multi-channel multi-modality 3DCNN framework for ASL recognition. The multiple channels contain RGB, Depth, and Optical flow while the multiple modalities include hand gestures, facial expressions and body poses. While the full size image is used to represent body pose, to better model hand gestures and the facial expression changes, the regions of hands and face are obtained from the RGB image based on the location guided by skeleton joints. The whole framework consists of two main components: proxy video generation and 3DCNN modeling. First, proxy videos are generated for each ASL sign by selecting a subset of frames spanning the whole video clip of each ASL sign, to represent the overall temporal dynamics. Then the generated proxy videos of RGB, Depth, Optical flow, RGB of hands, and RGB of face are fed into the multi-stream 3DCNN component. The predictions of these networks are weighted to obtain the final results of ASL recognition.}
\label{fig:pipeline}
\end{center}
\end{figure*}


\section{The Proposed Method  for ASL Recognition}

The pipeline of our proposed method is illustrated in Fig. \ref{fig:pipeline}. There are two main components in the framework: random temporal augmentation to generate proxy videos (which are representative of the overall temporal dynamics of video clip of an ASL sign) and 3DCNN to recognize the class label of the sign. 


\subsection{Random Temporal Augmentation for Proxy Video Generation}

The performance of the deep neural network greatly depends on the amount of the training data. Large-scale training data and different data augmentation techniques usually are needed for deep networks to avoid over-fitting. During training, different kinds of data augmentation techniques, such as random resizing and random cropping of images, are already widely applied in 3DCNN training. In order to capture the overall temporal dynamics, we apply a random temporal augmentation, to generate a proxy video for each sign video clip channel, by selecting a subset of frames, which has proved to be very effective for our proposed framework.

Videos are often redundant in the temporal dimension, and some consecutive frames are very similar without observable difference, as shown in Fig. \ref{fig:data_aug} (a) which displays $8$ consecutive frames in a video clip of an ASL sign while the proxy video in \ref{fig:data_aug} (b) displays the $8$ frames selected from the same video clip by random temporal augmentation. With the same number of frames, the proxy video provides more temporal dynamics. Thus,  proxy videos are generated to represent the overall temporal dynamics for each ASL word.

The process of proxy video generation by randomly sampling is formulated in Eq. (\ref{eq.randomly}) below: 

\begin{equation}
\label{eq.randomly}
S_{i} = random(\lfloor N/T \rfloor) + \lfloor N/T \rfloor *i,
\end{equation} 
 where $N$ is the number of frames of a sign video, $T$ is the number of the sampled frames from the video, $S_{i}$ is the $i$th sampled frame, $random(N/T)$ generates one random number ranging  $\lfloor 0, N/T \rfloor$ for every $i$. To generate the proxy video, each video is uniformly divided into $T$ intervals, and one frame is randomly sampled from every interval. If the total number of frames for a video is less than $T$, it is padded with the last frame to the length of $T$. These proxy videos make it feasible to train deep neural network on the proposed dataset.

\begin{figure*}[t]
\begin{center}
\includegraphics[width=\textwidth]{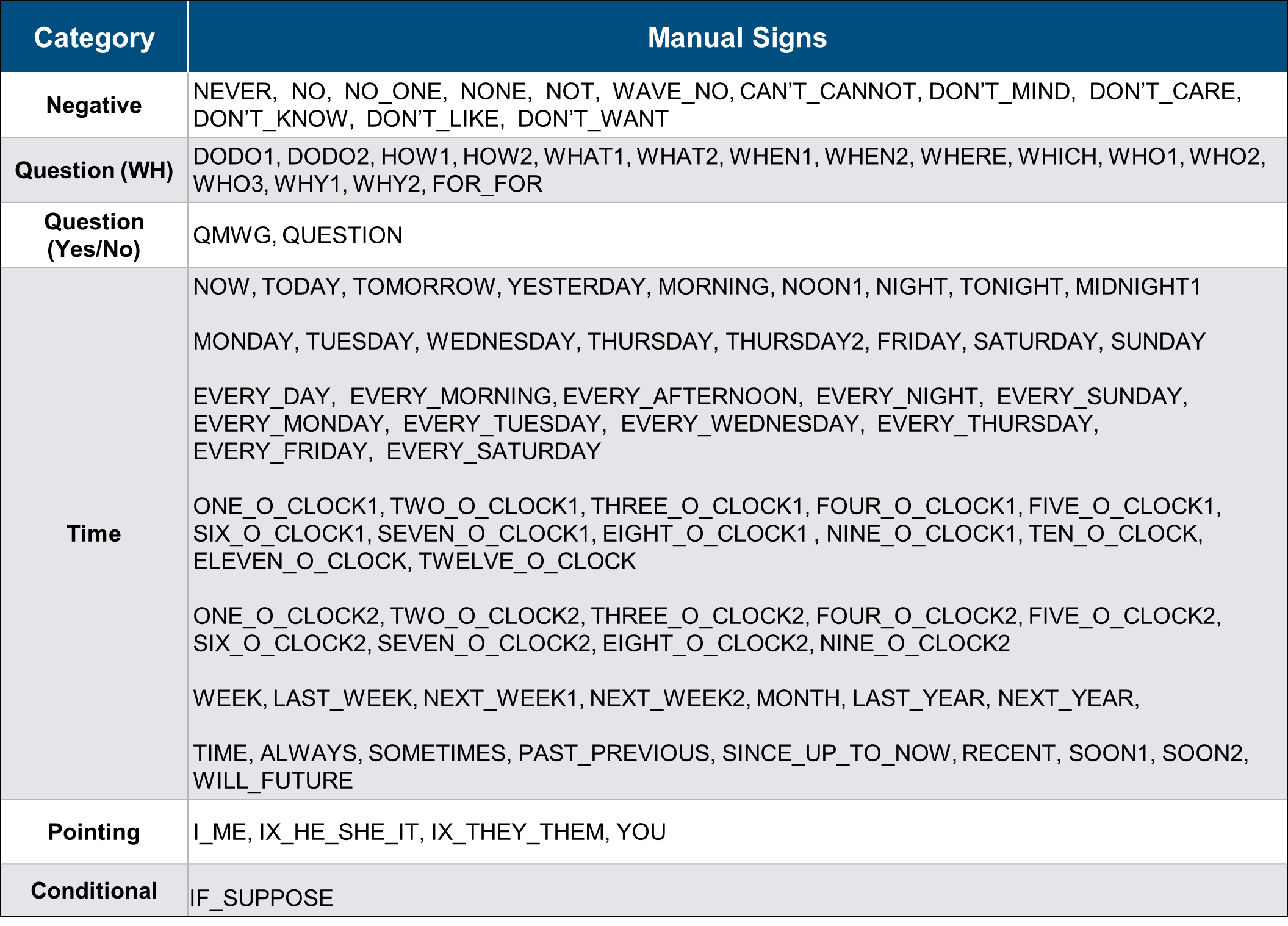}
\caption{The full list of the 100 ASL words in our ``ASL-100-RGBD" dataset under $6$ semantic categories. These ASL words have been strategically selected to support sign recognition technology for ASL education tools (many of these words consist of both hand gestures and facial expression changes.)}
\end{center}
\label{fig:WWVideo}
\end{figure*}

\subsection{3D Convolutional Neural Network}

3DCNN was first proposed for video action recognition \cite{3D}, and was improved in C3D \cite{C3D} by using a similar architecture to VGG \cite{VGG}. It  obtained the state-of-the-art performance for several video recognition tasks. The difference between the 2DCNN and 3DCNN operation is that 3DCNN has an extra  temporal dimension, which can capture the spatial and temporal information between video frames more effectively.

After the emergence of C3D, many 3DCNN models were proposed for video action recognition \cite{I3D},\cite{T3D},\cite{P3D}. 3D-ResNet was the 3D version of ResNet which introduced identical mapping to avoid gradient vanishing and explosion, which makes the training of very deep of convolutional neural networks feasible. Compared to 2D ResNet, the size of the convolution kernel is $\mathit{w \times h \times t}$ ($w$ is the width of the kernel, $h$ is the height of the kernel and $t$ is the temporal dimension of the kernel) in 3D-ResNet, while it is $\mathit{w \times h}$ in 2D-ResNet. In this paper, 3D-ResNet is chosen as the base network for ASL recognition. 

 In order to handle the three important elements of ASL recognition (hand gesture, facial expression, and body pose), a hybrid framework is designed including two 3DCNN networks: one for full body, to capture the full body movements including hands and face with the inputs of the multi-channel proxy videos generated from the full images including RGB, depth, and optical flow; and another for hand and face, to capture the coordinates of hands and face with the inputs of the multi-channel proxy videos generated from the cropped regions of left hand, right hand, and face. Note for the Hand-Face network, RGB and depth channels are employed for hand regions. The optical flow is not employed since it cannot accurately track the quick and large hand motions. For the face regions, only RGB channel is employed since facial expressions generally change much less in depth. The prediction results of the networks are weighted to obtain the final prediction of each ASL sign. 

 The optical flow images are calculated by stacking the $x$-component, the $y$-component, and the magnitude of the flow. Each value in the image is then rescaled to $\mathit{0}$ and $\mathit{255}$. This practice has yielded good performance in other studies \cite{donahue2014long, ng2015beyond}. As observed in the experimental results, by fusing all the features generated by RGB, optical flow, and depth images, the performance can be improved, which indicates that complementary information are provided by different channels in training deep neural networks.

\section{Proposed ASL Dataset: ``ASL-100-RGBD"}

As mentioned in Section 2.3, a new dataset has been collected for this research in collaboration with ASL computational linguistic researchers, from native ASL signers (individuals who have been using the language since very early childhood) who performed a word list of $\mathit{100}$ ASL signs (See the full list of ASL words in Fig. 4) by using a Kinect V2 camera. Participants responded affirmatively to the following screening question: Did you use ASL at home growing up, or did you attend a school as a very young child where you used ASL? Participants were provided with a slide-show presentation that asked them to perform a sequence of $\mathit{100}$ individual ASL signs, without lowering their hands between words.  Since this new dataset includes $\mathit{100}$ signs with RGB and depth data, we refer to it as the ``ASL-100-RGBD" dataset. 

During the recording session, a native ASL signer met the participant and conducted the session: prior research in ASL computational linguistics has emphasized the importance of having only native signers present when recording ASL videos so that the signer does not produce English-influenced signing \cite{lu2012cuny}. Several videos were recorded of each of the $\mathit{15}$ people, while they signed the $\mathit{100}$ ASL signs. Typically three videos were recorded from each person, to produce a total collection of $42$ videos (each video contains all the $100$ signs) and $\mathit{4,200}$ samples of ASL signs.

To facilitate this collection process, we have developed a recording system based on Kinect 2.0 RGB-D camera to capture multimodality (facial expressions, hand gestures, and body poses) from multiple channels of information (RGB video and depth video) for ASL recognition. The recordings also include skeleton and HDface information. The video resolution is $\mathit{1920}$ x $\mathit{1080}$ pixels for the RGB channel and $\mathit{512}$ x $\mathit{424}$ pixels for the depth channel respectively. 

The $\mathit{100}$ ASL signs in this collection were selected to strategically support research on sign recognition for ASL education applications, and the words were chosen based on vocabulary that is  traditionally included in introductory ASL courses.  Specifically, as discussed in \cite{huenerfauth2017evaluation}, our recognition system must identify a subset of ASL words that relate to a list of errors often made by students who are learning ASL.  Our proposed educational tool \cite{huenerfauth2017evaluation} would receive as input a video of a student who is performing ASL sentences, and the system would automatically identify whether the student's performance may include one of several dozen errors, which are common among students learning ASL. As part of the operation of this system, we require a sign-recognition component that can identify when a video of a person includes any of these $\mathit{100}$ words during the video (and the time period of the video when this word occurs).  When one of these $100$ key words are identified, then the system will consider other properties of the signer's movements \cite{huenerfauth2017evaluation}, to determine whether the signer may have made a mistake in their signing.

For instance, the $\mathit{100}$ ASL signs includes words related to questions (e.g. WHO, WHAT), time-phrases (e.g. TODAY, YESTERDAY), negation (e.g. NOT, NEVER), and other categories of words that relate to key grammar rules of ASL. A full listing of the words included in this dataset appears in Fig. 4. Note that there is no one-to-one mapping between English words and ASL signs, and some ASL signs have variations in their performance, e.g. due to geographic/regional differences or other factors.  For this reason, some words in Fig. 4 appear with integers after their name, e.g. THURSDAY and THURSDAY2, to reflect more than one variation in how the ASL word may be produced.  For instance, THURSDAY indicates a sign produced by the signer's dominant hand in the "H" alphabet-letter handshape gentle circling in space; whereas, THURSDAY2 indicates a sign produced with the signer's dominant hand quickly switching from the alphabet-letter handshape of "T" to "H" while held in space in front of the torso.  Both are commonly used ASL signs for the concept of "Thursday"; they simply represent two different ASL words that could be used for the same concept. 

As shown in Fig. 4, the words are grouped into $6$ semantic categories (Negative, WH Questions, Yes/No Questions, Time, Pointing, and Conditional), which in some cases suggest particular facial expressions that are likely to co-occur with these words when they are used in ASL sentences.  For instance, time-related phrases that appear at the beginning of ASL sentences tend to co-occur with a specific facial expression (head tilted back slightly and to the side, with eyebrows raised). Additional details about how detecting words in these various categories would be useful in the context of educational software appear in \cite{huenerfauth2017evaluation}.

After the videos were collected from participants, the videos were analyzed by a team of ASL linguists, who produced time-coded annotations for each video. The linguists used a coding scheme in which an English identifier label was used to correspond to each of the ASL words used in the videos, in a consistent manner across the videos. For example, all of the time spans in the videos when the human performed the ASL word ``NOT'' were labeled with the English string "NOT" in our linguistic annotation. 

Fig. 5 demonstrates several frames of each channel of an ASL sign from our dataset including RGB, skeleton joints ($25$ joints for every frame), depth map, basic face features (5 main face components), and HDFace (1,347 points). With the publication of this article, ASL-100-RGBD dataset\footnote{Some example videos can be found at our research website {\color{blue}http://media-lab.ccny.cuny.edu/wordpress/datecode/}\\}
will be released to the research community. 

\begin{figure}[t]
\begin{center}
\includegraphics[width=0.5\textwidth]{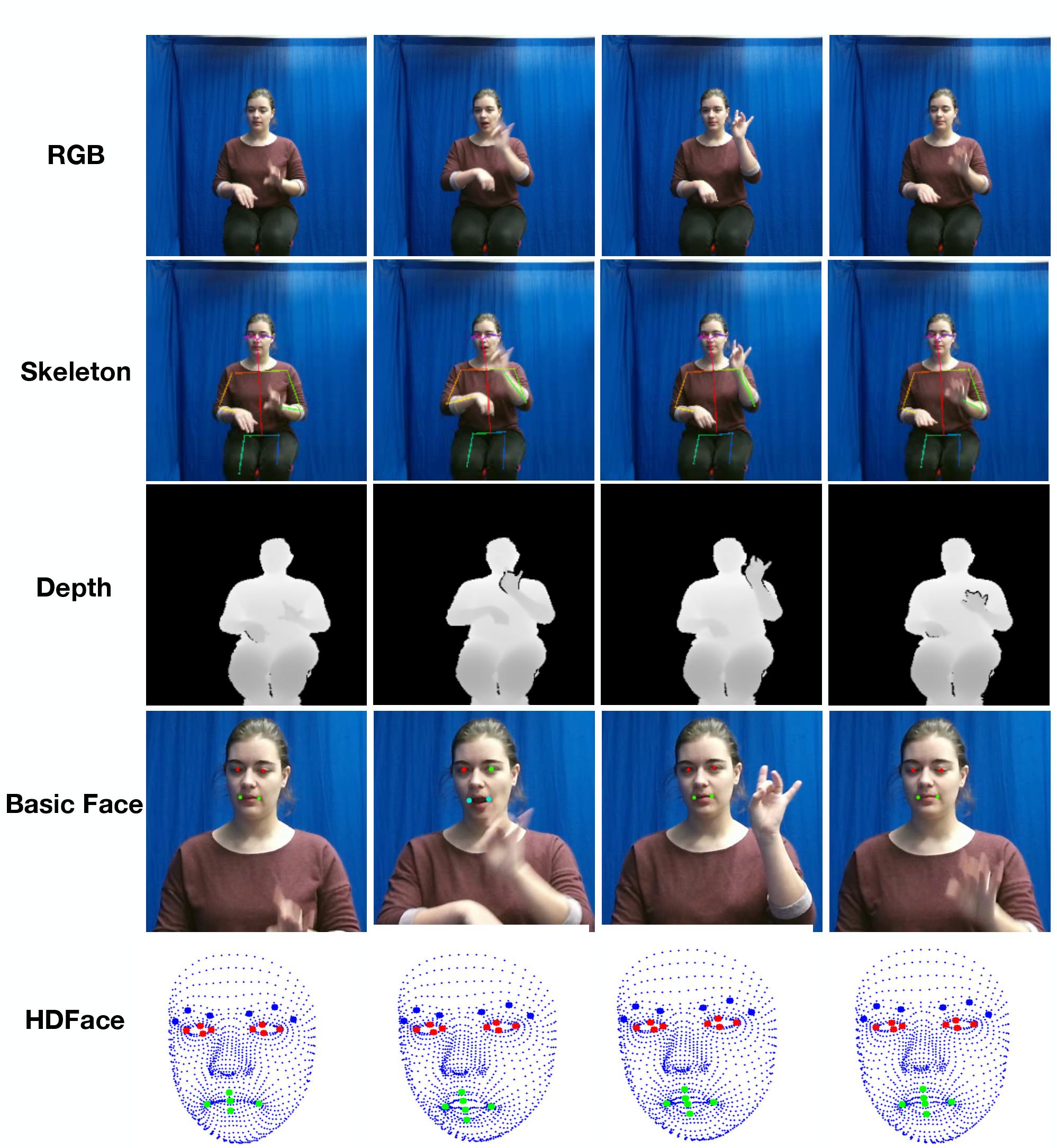}
\end{center}
   \caption{Four sample frames of each channel of an ASL sign from our dataset including RGB, skeleton joints ($25$ joints for every frame), depth map,  basic face features (5 main face components), and HDFace (1,347 points.)}
\label{fig:ASL100dataset}
\end{figure}

\section{ Experiments and Discussions}

In this section, extensive experiments are conducted to evaluate the proposed approach on the newly collected ``ASL-100-RGBD" dataset and the Chalearn LAP IsoGD dataset \cite{LargeScaleRGBDHandGesture16}.

\subsection{Implementation Details} 

Same 3D-ResNet architecture is employed for all experiments. Different channels and modalities are fed to the network as input. The input channels are RGB, Depth, RGBflow (i.e. Optical flow of RGB images), and Depthflow (i.e. Optical flow of depth images) of modalities including hands, face, and full body. The fusion of different channels are studied and compared.

Our proposed model is trained in PyTorch on four Titan X GPUs. To avoid over-fitting, the pretrained model from Kinetics or Chalearn dataset is employed and the following data augmentation techniques were used: random cropping (using a patch size of $112\times112$) and random rotation (with a random number of degrees in a range of [$-10$, $10$]). The models are then fine-tuned for $\mathit{50}$ epochs with an initial learning rate of $\lambda = 3\times10^{-3}$, reduced by a factor of $\mathit{10}$ after every $\mathit{25}$ epochs. 

To apply the pre-trained 3D-ResNet model on 3 bands in RGB image format to one channel depth images or optical flow images, the depth images are simply converted to 3 bands as RGB image format. For the optical flow images, the pre-trained 3D-ResNet model takes the $x$-component, the $y$-component, and the magnitude of flow as the R, G, and B bands in the RGB format. 


\subsection{Experiments on ASL-100-RGBD}


To prepare the training and testing for evaluation of the proposed method on ``ASL-100-RGBD" dataset, we first extracted the video clips for each ASL sign. We use $3,250$ ASL clips for training ( $\mathit{75\%}$ of the data) and the remaining $\mathit{25\%}$ ASL clips for testing. To ensure a subject independent evaluation, there is no same signer appearing in both training and testing datasets. To augment the data, a new 16-frame proxy video is generated from each video by selecting different subset of frames for each epoch during the training phase.  

\subsubsection{Effects of Data Augmentations}

The training dataset which contains $3,250$ ASL video clips of $100$ ASL manual signs is relatively small for 3DCNN training  and could easily cause an over-fitting problem. In order to extract more representative temporal dynamics as well as avoid over-fitting,  a random temporal augmentation technique is applied to generate proxy videos (a new proxy video for each epoch) for each ASL clip. The ASL recognition results of using the proposed proxy video (16 frames per video) are compared with the traditional method (using the same number of consecutive frames). The network, 3DResNet-34, dose not converge when trained with $16$ consecutive frames, while the network trained with proxy video obtained $68.4$\% on the testing dataset. This is likely due to the majority of movements being from hands in these videos and the consecutive frames could not effectively represent the temporal and spatial information. Therefore, the network could not distinguish the clips based on only $16$ consecutive frames. We also evaluate the effect of random cropping (using a patch size of $112\times112$) and random rotation (with a random number of degrees in a range of [$-10$, $10$]).

Table~\ref{table:augmentation} lists the effects of different data augmentation techniques for the performance for recognizing $100$ ASL words on only RGB channel. With the proxy videos, the 3DCNN model obtains $68.4$\% accuracy on the testing data for recognizing $100$ ASL signs. By adding the random crop, the performance is improved by $4.4$\% and adding the random rotation further improved the performance to $75.9$\%. In the following experiments, proxy videos together with random crop and random rotation are employed to augment the data.

\begin{table}[!htb]
\centering
\caption{The comparison of the performance of different data augmentation methods on only RGB channel with 16 frames for recognizing $100$ ASL manual signs. All the models are pretrained on Kinetics and finetuned on ASL-100-RGBD dataset. The best performance is achieved with the random proxy video, random crop, and random rotation.}
\resizebox{0.48\textwidth}{!}{
\begin{tabular}{l c c c c}
 \hline
  Augmentations & \multicolumn{4}{c}{Fusions}\\ 
 \hline
  \hline
  Random Proxy Video & \cellcolor{gray!70} \xmark & \cellcolor{gray!25}$\surd$& \cellcolor{gray!70}$\surd$ & \cellcolor{gray!25} $\surd$ \\
    
   Random Crop & \cellcolor{gray!70} \xmark & \cellcolor{gray!25} &\cellcolor{gray!70} $\surd$&\cellcolor{gray!25}$\surd$\\
   
   Random Rotation & \cellcolor{gray!70} \xmark & \cellcolor{gray!25}& \cellcolor{gray!70} & \cellcolor{gray!25}$\surd$\\
   
   \hline
   Performance & \cellcolor{gray!70} Not converging &\cellcolor{gray!25}68.4\% & \cellcolor{gray!70}72.8\% &\cellcolor{gray!25}75.9\%\\
  \hline
\end{tabular}}
\label{table:augmentation}
\end{table}

\vspace{-0.2in}

\subsubsection{Effects of Network Architectures}

In this experiment, the ASL recognition results of different number of layers at $18$, $34$, $50$, and $101$ for 3DResNet are compared on full RGB, optical flow, and depth images. As shown in Table~\ref{tab:architecture}, the performance of 3DResNet-18, 3DResNet-50, and 3DResNet-101 achieve comparable results on RGB channel. However, the performance on optical flow and depth channels are much lower than that of RGB channel because the network has been pre-trained on from Kinetics dataset which contains only RGB images. As shown in Table~\ref{tab:architecture}, 3DResNet-34 obtained the best performance for all RGB, optical flow, and depth channels. Hence, 3DResNet-34 is chosen for all the subsequent experiments. 

\begin{table}[!htb]
\centering
\caption{The effects of number of layers for 3DResNet with $16$ frames on RGB, optical flow, and depth channels. All the models are pre-trained on Kinetics and finetuned on ASL-100-RGBD dataset.}
\resizebox{0.45\textwidth}{!}{
\begin{tabular}{l c c c }
  \hline
  Network   &  RGB  (\%) & Optical Flow (\%) & Depth (\%)\\
  \hline
  \hline
  3DResNet-18    & \cellcolor{gray!70} $73.2$     & \cellcolor{gray!25} $61.9$  & \cellcolor{gray!70}$65.0$\\
  
  \textbf{3DResNet-34}    &\cellcolor{gray!70}\bm{$75.9$}     &\cellcolor{gray!25}\bm{$62.8$}  &\cellcolor{gray!70}\bm{$66.5$}  \\
  
  3DResNet-50    &\cellcolor{gray!70}$72.3$     &\cellcolor{gray!25}$55.4$  &\cellcolor{gray!70}$62.0$\\
  
  3DResNet-101   &\cellcolor{gray!70}$72.5$     &\cellcolor{gray!25}$55.0$  &\cellcolor{gray!70}$61.5$\\
  \hline
\end{tabular}}
\label{tab:architecture}
\end{table}


\subsubsection{Effects of Pre-trained Models}

\begin{figure}[t]
\begin{center}
\includegraphics[width=0.5\textwidth]{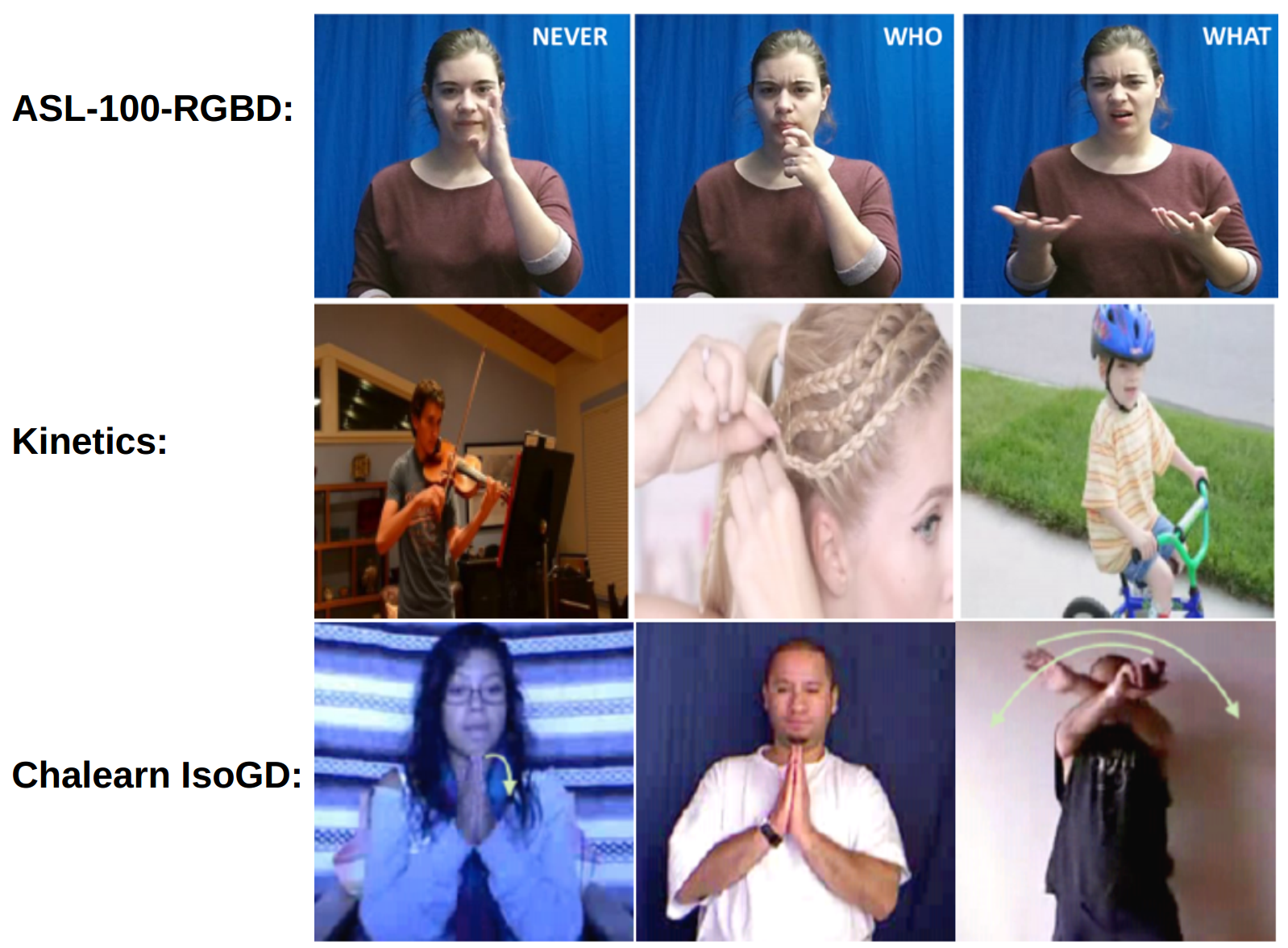}
\end{center}
   \caption{Example images of three datasets. ASL-100-RGBD: various ASL signs. Kinetics dataset: consisting of diverse human actions, involving different parts of body. Chalearn IsoGD: various hands gestures including mudras (Hindu/ Buddhist hand gestures) and diving signals.}
\label{fig:dataset}
\end{figure}

To evaluate the effects of pre-trained models, we fine-tune 3DResNet-34 with pretrained models from the Kinectics \cite{Kinetics} and the Chalearn LAP IsoGD datasets \cite{LargeScaleRGBDHandGesture16}, respectively. Kinetics dataset consists of RGB videos of diverse human actions which involve different parts of body while the Chalearn LAP IsoGD dataset contains both RGB and depth videos of various hand gestures including mudras (Hindu/ Buddhist hand gestures), Chinese numbers and diving signals, as shown in Fig. \ref{fig:dataset}.

The results are shown in Table \ref{tab:comp_chalearn_kinetics}. The temporal duration is fixed to $16$ and the channels are RGB, Depth, and RGBflow. In all channels, the performance using the pretrained models from Chalearn dataset is better than pretrained models from Kinetics dataset. This is probably because all the videos in Chalearn dataset are focused on hand gestures and the network trained on this dataset can learn prior knowledge of hand gestures. The Kinetics dataset consists of general videos from YouTube and the network focuses on the prior knowledge of motions. Therefore, for each channel the pretrained model on the same channel of Chalearn dataset is used in the subsequent experiments.

\begin{table}[!htb]
\centering
\caption{The comparison of the performance of recognizing $100$ ASL words on 3DResNet-34 with different pretrained models.}
\resizebox{0.35\textwidth}{!}{
\begin{tabular}{l c c}
  \hline
  Channels  &  Kinetics (\%) &  Chalearn (\%) \\
  \hline
  \hline
  RGB    & \cellcolor{gray!70} $75.9$     & \cellcolor{gray!25} $76.38$   \\
  
  Depth   &\cellcolor{gray!70} $66.5$     &\cellcolor{gray!25} $68.18$  \\
  
  RGB Flow   &\cellcolor{gray!70} $62.8$    &\cellcolor{gray!25} $66.79$ \\
  \hline
\end{tabular}}
\label{tab:comp_chalearn_kinetics}
\end{table}

\vspace{-0.2in}

\subsubsection{Effects of Temporal Duration of Proxy Videos }

We study the effects of temporal duration (i.e. number of frames used in proxy videos) by finetuning  3DResNet-34 on ASL-100-RGBD dataset at different temporal duration in proxy videos at 16, 32, and 64 respectively. Note that the same temporal duration is also used to train the corresponding pre-trained model on the Chalearn dataset. Results are shown in Table \ref{tab:asl_diff_temp}. The performance of the network with $64$ frames achieves the best performance. Therefore, 3D-ResNet-34 with $64$ frames is used in all the following experiments.

\begin{table}[!htb]
\centering
\caption{The comparison of the performance of networks with different temporal duration (i.e. number of frames used in proxy videos). All the models are pretrained on Chalearn dataset and finetuned on ASL-100-RGBD dataset by using same temporal duration.}
\resizebox{0.48\textwidth}{!}{
\begin{tabular}{ l c c c }
  \hline
 Channel &  16 frames (\%) & 32 frames (\%) & 64 frames (\%)\\
  \hline
  \hline
 RGB  & \cellcolor{gray!70} $76.38$ & \cellcolor{gray!25} $80.73$ & \cellcolor{gray!70} $87.83$ \\   

 Depth & \cellcolor{gray!70} $68.18$ & \cellcolor{gray!25} $74.21$ & \cellcolor{gray!70} $81.93$ \\   
 
 RGB Flow & \cellcolor{gray!70} $66.79$ & \cellcolor{gray!25} $71.74$ & \cellcolor{gray!70} $80.51$ \\  
  \hline 
\end{tabular}}
\label{tab:asl_diff_temp}
\end{table}

\vspace{-0.2in}

\subsubsection{Effects of Different Input Channels}

In this section, we examine the fusion results of different input channels. The RGB channel provides global spatial and temporal appearance information, the depth channel provides the distance information, and the optical flow channel captures the motion information. The network is finetuned on the three input channels respectively. The average fusion is obtained by weighting the predicted results.

Table~\ref{table:input_fusion} shows the performance of ASL recognition on ASL-100-RGBD dataset for each input channel and different fusions. While RGB channel alone achieves $\mathit{87.83\%}$, by fusing with optical flow, the performance is boosted up to $\mathit{89.02\%}$. With the fusion of all the three channels (RGB, Optical flow, and Depth), the performance is further improved to $\mathit{89.91\%}$. This indicates that depth and optical flow channels contain complementary information to RGB channel for ASL recognition.

\begin{table}[!htb]
\centering
\caption{The comparison of the performance of networks with different input channels and their fusions. All the models are pretrained on Chalearn dataset and finetuned on ASL-100-RGBD dataset with $64$ frames.} 
\resizebox{0.48\textwidth}{!}{
\begin{tabular}{l c c c c c c}
 \hline
  Channels & \multicolumn{6}{c}{Fusions}\\ 
 \hline
  \hline
   RGB & \cellcolor{gray!25}$\surd$ &\cellcolor{gray!70} & \cellcolor{gray!25}& \cellcolor{gray!70}$\surd$&\cellcolor{gray!25}$\surd$& \cellcolor{gray!70}$\surd$\\
   
   Depth & \cellcolor{gray!25} &\cellcolor{gray!70} $\surd$ & \cellcolor{gray!25}& \cellcolor{gray!70}$\surd$&\cellcolor{gray!25}& \cellcolor{gray!70}$\surd$\\
   Optical Flow & \cellcolor{gray!25}& \cellcolor{gray!70} &\cellcolor{gray!25}$\surd$ &\cellcolor{gray!70} & \cellcolor{gray!25}$\surd$& \cellcolor{gray!70}$\surd$\\
    \hline
   Performance &\cellcolor{gray!25}87.83\% & \cellcolor{gray!70}81.93\% & \cellcolor{gray!25}80.51\%&\cellcolor{gray!70}89.91\% &\cellcolor{gray!25}89.02\% & \cellcolor{gray!70}89.71\\
  \hline
\end{tabular}}
\label{table:input_fusion}
\end{table}

\subsubsection{Effects of Different Modalities}

\begin{figure}[!htp]
\begin{center}
\includegraphics[width=0.5\textwidth]{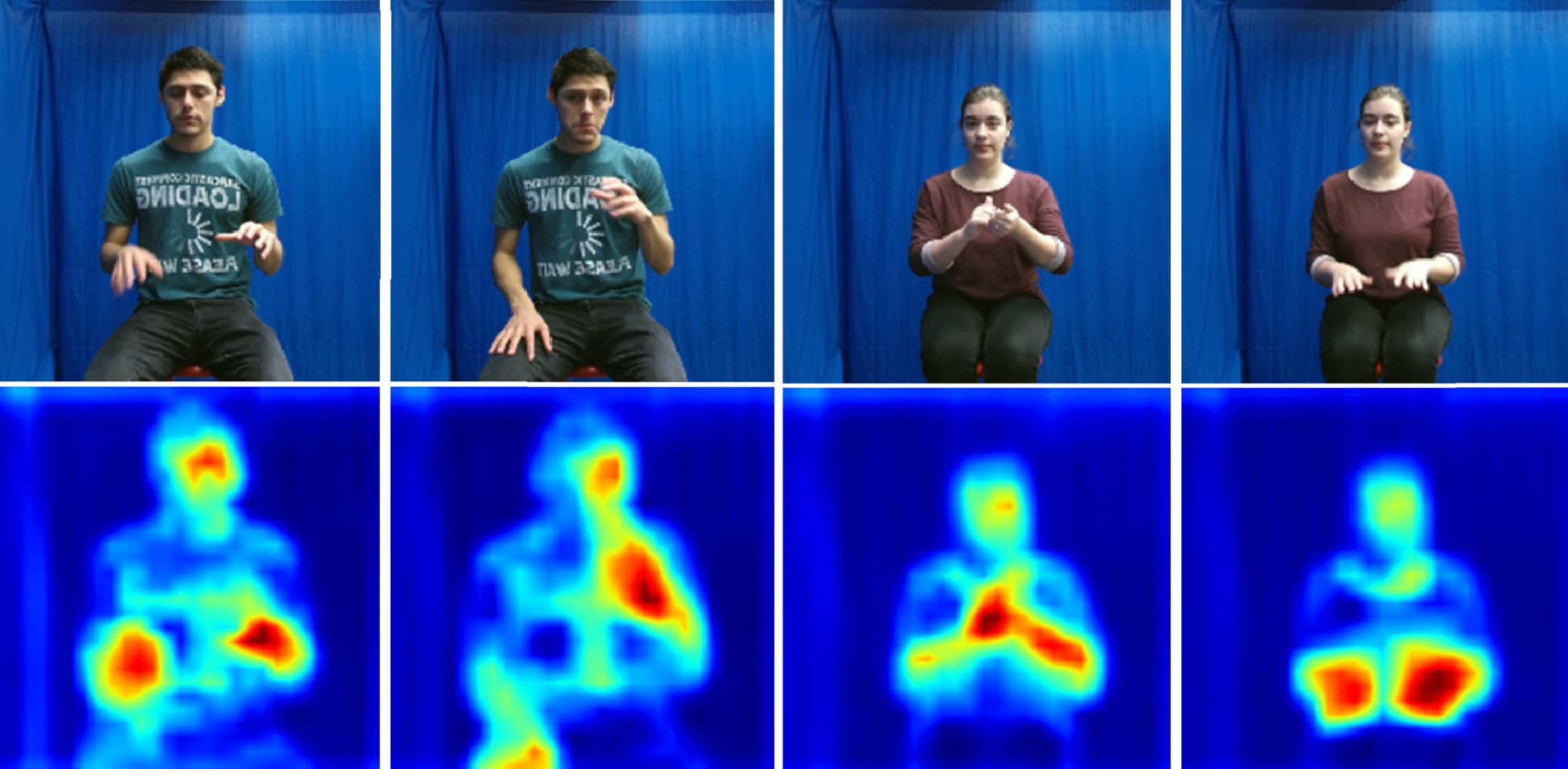}
\end{center}
   \caption{The example RGB images and their corresponding attention maps from the fifth convolution layer of the 3DResNet-34 on our test dataset of ASL recognition which the hands and face have most of the attention.
   }
\label{fig:feature}
\end{figure}

We attain further insight into the learned features of the 3DCNN model for RGB channel. In Fig~\ref{fig:feature} we visualize some examples of the attention maps of the fifth convolution layer on our test dataset generated by the trained RGB 3DCNN model for ASL recognition. These attention maps are computed by averaging the magnitude of activations of convolution layer which reflect the attention of the network. The attention maps show that the model mostly focused on \textbf{hands} and \textbf{face} of the signer during the ASL recognition process. 

Hence, we conduct experiments to analyze the effects of each modality (hand gestures, facial expression, and body poses) with the RGB channel. As shown in Fig. \ref{fig:pipeline}, the hand regions and the face regions are obtained from the RGB image based on the location guided by skeleton joints. The performance of each modality and their fusions are summarized in Table ~\ref{table:multi-madality}. 

\begin{table}[!htb]
\centering
\caption{The comparison of the performance of different modalities and their fusions. All the models are pretrained on Chalearn dataset and finetuned on ASL-100-RGBD dataset with $64$ frames.} 
\resizebox{0.45\textwidth}{!}{
\begin{tabular}{l c c c c}
 \hline
  Channels & \multicolumn{4}{c}{Fusions}\\ 
 \hline
  \hline
   Body & \cellcolor{gray!25}$\surd$ & \cellcolor{gray!70} &\cellcolor{gray!25}$\surd$ &\cellcolor{gray!70}$\surd$\\
   Hand & \cellcolor{gray!25} &\cellcolor{gray!70} $\surd$& \cellcolor{gray!25}$\surd$& \cellcolor{gray!70}$\surd$\\
   Face & \cellcolor{gray!25} &\cellcolor{gray!70} & \cellcolor{gray!25} & \cellcolor{gray!70}$\surd$\\
       \hline
   Performance &\cellcolor{gray!25}$87.83$\% & \cellcolor{gray!70}$80.9$\% & \cellcolor{gray!25}$89.81$\%&\cellcolor{gray!70}$91.5$\% \\
  \hline
\end{tabular}}
\label{table:multi-madality}
\end{table}

In addition to the accuracy of ASL sign recognition, we further analyzed the accuracy of the six categories (see Fig. 4 for details) for each modality and their combinations in Table \ref{table:category_multi-madality}. For the categories that involve many facial expressions, such as \textbf{Question(Yes/No)} and \textbf{Negative}, the accuracy of hand modality is improved by more than $15$\% after fusion with face modality. For the \textbf{Conditional} category which utilizes more subtle facial expressions, the accuracy of hand modality is not improved after fusion with face modality. 

\vspace{-0.2in}

\begin{table}[h]%
\footnotesize
\small
\caption{The performance (\%) of different modalities and their fusions on six categories listed in Fig. 4:  Conditional (Cond), Negative (Neg), Pointing (Point), Question (WH), Yes/No Question (Y/N) and Time. The last column is the accuracy (\%) for ASL signs.}
\resizebox{0.42\textwidth}{!}{\begin{minipage}{0.5\textwidth}
\begin{center}
\begin{tabular}{l c c c c c c c}
  \hline
  Modalities & Cond  &Neg &Point &WH &Y/N  &Time & Acc  \\
  \hline
  \hline
  Hand           
  &\cellcolor{gray!70}$90.0$ &\cellcolor{gray!25}$78.1$ &\cellcolor{gray!70}$68.4$ &\cellcolor{gray!25}$84.3$ &\cellcolor{gray!70}$68.4$ &\cellcolor{gray!25}$81.4$ &\cellcolor{gray!70}$80.9$\\

  Body     
  &\cellcolor{gray!70}$100.0$ &\cellcolor{gray!25}$87.4$ &\cellcolor{gray!70}$84.2$ &\cellcolor{gray!25}$88.0$ &\cellcolor{gray!70}$89.5$ &\cellcolor{gray!25}$87.6$ &\cellcolor{gray!70}$87.83$ \\
 
  Body+Hand     
  &\cellcolor{gray!70}$90.9$ &\cellcolor{gray!25}$86.6$ &\cellcolor{gray!70}$89.5$ &\cellcolor{gray!25}$88.7$ &\cellcolor{gray!70}$94.7$ &\cellcolor{gray!25}$90.2$ &\cellcolor{gray!70}$89.81$\\
 
 Body+Hand+Face 
 &\cellcolor{gray!70}$90.9$ &\cellcolor{gray!25}$93.3$ &\cellcolor{gray!70}$84.2$ &\cellcolor{gray!25}$90.6$ &\cellcolor{gray!70}$84.2$ &\cellcolor{gray!25}$91.8$ &\cellcolor{gray!70}$91.5$\\
  \hline
\end{tabular}
\end{center}
\end{minipage}}
\label{table:category_multi-madality}
\end{table}%

\vspace{-0.1in}

\subsubsection{Fusions of Different Channels and Modalities}

The fusion results of different input channels and modalities on ASL-100-RGBD dataset are shown in Table \ref{table:final_asl_res}. The experiments are based on 3DResNet-34 with $64$ frames, pretrained on Chalearn dataset. Among all the models, fusion of \textbf{RGB+Depth+Hands RGB+ Face RGB} achieves the best performance with $92.88$\% accuracy. Adding RGBflow to this combination results in $92.48$\% accuracy which is comparable but not improved since the channels have redundant information.

\begin{table}[!htb]
\centering
\caption{Performance of 3DResNet-34 with 64 from fusion of different channels and modalities.} 
\resizebox{0.45\textwidth}{!}{
\begin{tabular}{l c c c c}
 \hline
Channels & \multicolumn{4}{c}{Fusions}\\ 
\hline
\hline
RGB & \cellcolor{gray!25} $\surd$ & \cellcolor{gray!70}$\surd$ &\cellcolor{gray!25} &\cellcolor{gray!70}$\surd$\\
Depth & \cellcolor{gray!25} $\surd$ & \cellcolor{gray!70} $\surd$ &\cellcolor{gray!25}$\surd$ &\cellcolor{gray!70}$\surd$\\
RGBflow & \cellcolor{gray!25} $\surd$ & \cellcolor{gray!70} $\surd$ &\cellcolor{gray!25}$\surd$ &\cellcolor{gray!70}\\
RGB of Hands & \cellcolor{gray!25} $\surd$ &\cellcolor{gray!70} $\surd$ & \cellcolor{gray!25}$\surd$& \cellcolor{gray!70}$\surd$\\
RGB of Face & \cellcolor{gray!25} &\cellcolor{gray!70} $\surd$ & \cellcolor{gray!25}$\surd$ & \cellcolor{gray!70}$\surd$\\
\hline
Performance &\cellcolor{gray!25} $91.19$\% & \cellcolor{gray!70} $92.48$\% & \cellcolor{gray!25} $92.48$\% &\cellcolor{gray!70} $92.88$\% \\
\hline
\end{tabular}}
\label{table:final_asl_res}
\end{table}

\vspace{-0.2in}

\subsection{Experiments on Chalearn LAP IsoGD dataset }

\subsubsection{Effects of Network Architectures}

The 3D-ResNet is pre-trained on Kinetics \cite{Kinetics} for all the experiments in this section. To find the best network architecture for Chalearn dataset, the parameters of 3D-ResNet are studied on RGB videos. The results are shown in Table \ref{tab:chalearn_diff_parameters_1}.
By changing the number of layers to $18$, $34$, $50$ while fixing the temporal duration to $32$, ResNet-34 achieved the best accuracy.

\begin{table}[!htb]%
\centering
\caption{Ablation study of number of layers of the network on RGB videos of Chalearn Dataset.}
\resizebox{0.35\textwidth}{!}{
\begin{tabular}{l c c}
  \hline
 Network & Temporal Duration & Accuracy \\
  \hline
  \hline
    ResNet-18 & \cellcolor{gray!70}$32$ &  \cellcolor{gray!25}$52.69\%$  \\
  
    \textbf{ResNet-34} & \cellcolor{gray!70}\bm{$32$} & \cellcolor{gray!25} \bm{$56.28\%$} \\
   
    ResNet-50  & \cellcolor{gray!70} $32$ & \cellcolor{gray!25} $54.57\%$ \\
  \hline
\end{tabular}}
\label{tab:chalearn_diff_parameters_1}
\end{table}

We also examined the performance of ResNet-34 by changing the temporal duration to $16$, $32$, and $64$. Our results indicate that ResNet-34 with $64$ frames has the best architecture for Chalearn dataset, as shown in Table \ref{tab:chalearn_diff_parameters}.

\begin{table}[!htb]%
\centering
\caption{Ablation study of temporal duration on RGB videos of Chalearn Dataset.}
\resizebox{0.35\textwidth}{!}{
\begin{tabular}{l c c}
  \hline
 Network & Temporal Duration & Accuracy \\
  \hline
  \hline
 ResNet-34 &\cellcolor{gray!70} $16$ &  \cellcolor{gray!25} $45.00\%$ \\

ResNet-34 & \cellcolor{gray!70} $32$ &  \cellcolor{gray!25} $56.28\%$ \\

 \textbf{ResNet-34} & \cellcolor{gray!70} \bm{$64$} &  \cellcolor{gray!25} \bm{$58.32\%$} \\
  \hline 
\end{tabular}}
\label{tab:chalearn_diff_parameters}
\end{table}

\subsubsection{Effects of Different Channels and Modalities}

We evaluate the effects of different channels including RGB, RGB flow, Depth, and Depth flow. Because the Chalearn dataset is designed for hand gesture recognition, we further analyze the effects of different hands (left and right), as well as the whole body. We develop a method to distinguish left and right hands in Chalearn Isolated Gesture dataset, and will release the coordinates of hands (distinguished between right and left hands) with the publication of this article. Since the Chalearn dataset is collected for recognizing hand gestures, here, the face channel is not employed.

We train $12$ 3D-ResNet-34 networks with $64$ frames by using different combinations of channels and modalities respectively and show the results in table \ref{tab:chalearn_all_channels}. The accuracy of right hand is significantly higher than the left hand. The reason is that for most of the gestures in Chalearn dataset, the right hand is dominant and the left hand does not move much for many hand gestures.

\begin{table}[!htb]%
\centering
\caption{Performance of 3D-ResNet-34 with $64$ frames on Chalearn Dataset for different channels and modalities.}
\resizebox{0.45\textwidth}{!}{
\begin{tabular}{l c c c}
  \hline
 Channel &  Global Channel (\%) & Left Hand (\%) & Right Hand (\%)\\
  \hline
  \hline
 RGB  & \cellcolor{gray!70} $58.32$ & \cellcolor{gray!25} $18.01$ & \cellcolor{gray!70} $48.58$ \\

 Depth & \cellcolor{gray!70} $63.16$ & \cellcolor{gray!25} $19.43$ & \cellcolor{gray!70} $54.15$ \\

 RGB Flow & \cellcolor{gray!70} $60.26$ & \cellcolor{gray!25} $21.97$ & \cellcolor{gray!70} $48.79$ \\

 Depth Flow & \cellcolor{gray!70} $55.37$ & \cellcolor{gray!25} $20.28$ & \cellcolor{gray!70} $47.07$\\ 
  \hline 
\end{tabular}}
\label{tab:chalearn_all_channels}
\end{table}

\vspace{-0.2in}

\subsubsection{Effects of Fusions on different channels and Modalities}

Here we analyze the effects of average fusion on different channels and modalities. The results are shown in Table \ref{tab:chalearn_fusion}. Using only  RGB and Depth channels, the accuracy is  $67.58$\% which is improved to $69.97$\% by adding RGB flow. We observe that among all different triplets of channels, \emph{Right Hand RGB + Depth + RGBflow} has the highest accuracy at $73.32$\%. By applying the average fusion on four channels \emph{RGB+ RGBflow+ Right Hand RGB + Right Hand Depth}, our model achieves the accuracy about \textbf{$75.88$\%} which outperforms the average fusion results of all previous work on Chalearn dataset. In the-state-of-the-art work of \cite{narayana2018gesture}, the accuracy of average fusion is $71.93$\% for $7$ channels and $70.37$\% for $12$ channels, respectively. 

Finally, the average fusion of all global channels (RGB, RGB flow, Depth, Depth flow) and Right hand channels ( Right hand RGB, Right hand RGB flow, Right hand Depth, Right hand Depth flow) resulted in $76.04$\% accuracy and the accuracy of 12 channels together resulted in $75.68$\%. This means that the 12 channels contain redundant information, and adding more channels does not necessarily improve the results.

\begin{table}[!htb]
\centering
\caption{Performance of 3DResNet-34 with $64$ frames for fusion of different channels and modalities on Charlearn dataset.} 
\resizebox{0.48\textwidth}{!}{
\begin{tabular}{l c c c c c}
 \hline
Channels & \multicolumn{5}{c}{Fusions}\\ 
\hline
\hline
RGB & \cellcolor{gray!25} $\surd$ & \cellcolor{gray!70} $\surd$ &\cellcolor{gray!25} &\cellcolor{gray!70} $\surd$  &\cellcolor{gray!25}$\surd$\\
Depth & \cellcolor{gray!25} $\surd$ & \cellcolor{gray!70} $\surd$ &\cellcolor{gray!25} $\surd$ &\cellcolor{gray!70} &\cellcolor{gray!25}$\surd$\\
RGBflow & \cellcolor{gray!25}  & \cellcolor{gray!70}$\surd$ &\cellcolor{gray!25} $\surd$&\cellcolor{gray!70}$\surd$ &\cellcolor{gray!25}$\surd$\\
RGB of Right Hand  & \cellcolor{gray!25}  & \cellcolor{gray!70} &\cellcolor{gray!25}$\surd$  &\cellcolor{gray!70}$\surd$  &\cellcolor{gray!25}$\surd$ \\
Depth of Right Hand  & \cellcolor{gray!25}  & \cellcolor{gray!70} &\cellcolor{gray!25} &\cellcolor{gray!70}$\surd$ &\cellcolor{gray!25}$\surd$\\
Performance &\cellcolor{gray!25} $67.58$\% & \cellcolor{gray!70} $69.97$\% & \cellcolor{gray!25} $73.32$\% &\cellcolor{gray!70} $75.53$\% &\cellcolor{gray!25}$75.88 $\%\\
\hline
\end{tabular}}
\label{tab:chalearn_fusion}
\end{table}

\vspace{-0.2in}

\subsubsection{Comparison with the-state-of-the-arts}

Our framework achieves accuracy of $75.88$\% and $76.04$\% from the fusion of $5$ and $8$ channels, respectively, on Chalearn IsoGD dataset. Table \ref{table:sota_results} lists the state-of-the-art results from Chalearn IsoGD competition 2017 as well as a recent paper, FOANet \cite{narayana2018gesture}. As shown in the table, in terms of the \textit{Average Fusion}, our framework achieves around $6$\% higher accuracy than the-state-of-the-arts methods. 

\begin{table}[!htb]%
\caption{Comparison with State-of-the-art Results on Chalearn IsoGD Dataset.}
\begin{minipage}{\columnwidth}
\begin{center}
\begin{tabular}{l|c}
\hline
Framework  & Accuracy on Test Set (\%) \\
\hline
\hline
\textbf{Our Results} & \bm{$76.04$} \\ 
\hline
FOANet (Average Fusion) \cite{narayana2018gesture}  & $70.37$ \\ 
\hline
Miao et al. (ASU)  \cite{miao2017multimodal}  & $67.71$  \\
\hline
SYSU-IEEE    & $67.02$  \\
\hline
Lostoy    &$65.97$  \\ 
\hline 
Wang et al. (AMRL) \cite{wang2017large}   &$65.59$  \\ 
\hline
Zhang et al. (XDETVP) \cite{zhang2017learning}  &$60.47$  \\ 
\hline
\end{tabular}
\end{center}
\label{table:sota_results}
\end{minipage}
\end{table}

It is worth noting that FOANet \cite{narayana2018gesture} reported the accuracy of $82.07$\% by applying \textit{Sparse Fusion} on the softmax scores of 12 channels (combinations of right hand, left hand, and whole body while each has 4 channels of  RGB, Depth, RGBflow and Depthflow). The purpose of using sparse fusion is to learn which channels are important for each gesture.  The accuracy of FOANet framework using average fusion is $70.37$\% which is around $6$\% lower than our results and nearly $12$\% lower than the accuracy of sparse fusion. While the authors of FOANet \cite{narayana2018gesture} had reported a $12$\% boost from using sparse fusion in their original experiments, our experiments do not reveal such a boost when implementing a system following the technical details provided in \cite{narayana2018gesture}.


Table \ref{tab:chalearn_compare} lists the accuracy on individual channels of our network and FOANet \cite{narayana2018gesture}. In this table, the values inside the parenthesis represent the accuracy of FOANet. As shown in the table, in the Global channel, our framework outperforms FOANet in all the four channels by $10$\% to $25$\%. Also, for the RGB of Right Hand, we obtain a comparable accuracy (~$48$\%) as FOANet. However, FOANet is outperforming our results in the Right Hand for Depth, RGBflow, and Depthflow by nearly $10$\%. From our experiments, the performance of "Global" channels (whole body) in general is superior to the Local channels (Right/ Left Hand) because the Global channels include more information. By using the similar architecture, FOANet reported $64$\% accuracy from Depth of Right Hand and $38$\% from Depth of the entire frame. Instead, our framework achieves more consistent results. For example, in our framework the accuracy of Depth channel is higher than RGB and RGBflow for both Global and Right Hand, while the accuracy  in FOANet for Depth and RGB are almost the same in the Global channel (around $40$\%) but very different in the Right Hand channel ($17$\% difference.)

\begin{table}[!htb]%
\centering
\caption{The accuracy ($\%$) of 12 channels on the test set of Chalearn IsoGD Dataset. Comparison between our framework and FOANet \cite{narayana2018gesture}. The values inside the parenthesis belong to FOANet.}
\resizebox{0.5\textwidth}{!}{
\begin{tabular}{ l c c c}
  \hline
 Channel &  Global Channel (\%) & Left Hand (\%) & Right Hand (\%) \\
  \hline
  \hline 
 RGB  &  \cellcolor{gray!25} \bm{$58.32$} ($41.27$) & \cellcolor{gray!70} \bm{$18.01$} ($16.63$)  & \cellcolor{gray!25} \bm{$48.58$} ($47.41$) \\

 Depth &  \cellcolor{gray!25} \bm{$63.16$} ($38.50$) & \cellcolor{gray!70} \bm{$19.43$} ($24.06$) & \cellcolor{gray!25} \bm{$54.15$} ($64.44$) \\

 RGB Flow & \cellcolor{gray!25} \bm{$60.26$} ($50.96$) &  \cellcolor{gray!70} \bm{$21.97$} ($24.02$) & \cellcolor{gray!25} \bm{$48.79$} ($59.69$) \\
 
 Depth Flow & \cellcolor{gray!25} \bm{$55.37$} ($42.02$) & \cellcolor{gray!70} \bm{$20.28$} ($22.71$) &  \cellcolor{gray!25} \bm{$47.07$} ($58.79$)\\ 
  \hline 
\end{tabular}}
\label{tab:chalearn_compare}
\end{table}

\vspace{-0.2in}

\section{Conclusion}
In this paper, we have proposed a 3DCNN-based multi-channel and multi-modality framework, which learns complementary information and embeds the temporal dynamics in videos to recognize manual signs of ASL from RGB-D videos. To validate our proposed method, we collaborate with ASL experts to collect an ASL dataset of 100 manual signs including both hand gestures and facial expressions with full annotation on the word labels and temporal boundaries (starting and ending points.) The experimental results have demonstrated that by fusing multiple channels in our proposed framework, the accuracy of recognizing ASL signs can be improved. This technology for identifying the appearance of specific ASL words has valuable applications for technologies that can benefit people who are DHH \cite{camgoz2017subunets, koller2016deep, koller2015deep, pigou2017gesture, kumar2018independent, palmeri2017sign, liu2015rgbd}. As an additional contribution, our ``ASL-100-RGBD" dataset, will be released to enable other members of the research community to use this resource for training or evaluation of models for ASL recognition. The effectiveness of the proposed framework is also evaluated  on the Chalearn IsoGD Dataset. Our method has achieved $75.88\%$ accuracy using only 5 channels which is $5.51$\% higher than the state-of-the-art work using 12 channels in terms of average fusion.

\begin{acknowledgements}
This material is based upon work supported by the National Science Foundation
under award numbers 1400802, 1400810, and 1462280.
\end{acknowledgements}

\bibliographystyle{spmpsci}      
\bibliography{Arxiv}

\begin{thebibliography}{10}
\providecommand{\url}[1]{{#1}}
\providecommand{\urlprefix}{URL }
\expandafter\ifx\csname urlstyle\endcsname\relax
  \providecommand{\doi}[1]{DOI~\discretionary{}{}{}#1}\else
  \providecommand{\doi}{DOI~\discretionary{}{}{}\begingroup
  \urlstyle{rm}\Url}\fi

\bibitem{Ncra}
American deaf and hard of hearing statistics.
\newblock https://www.nidcd.nih.gov/health/statistics/quick-statistics-hearing

\bibitem{IntelRealSense}
Intel realsense technology: Observe the world in 3d.
\newblock
  https://www.intel.com/content/www/us/en/architecture-and-technology/realsense-overview.html
   (2018)

\bibitem{OrbbecAstra}
Orbbec astra.
\newblock https://orbbec3d.com/product-astra/  (2018)

\bibitem{MSKinect2}
Set up kinect for windows v2 or an xbox kinect sensor with kinect adapter for
  windows.
\newblock
  https://support.xbox.com/en-US/xbox-on-windows/accessories/kinect-for-windows-v2-setup
   (2018)

\bibitem{vonAgris08}
von Agris, U., Knorr, M., Kraiss, K.F.: The significance of facial features for
  automatic sign language recognition.
\newblock In: Proceedings of IEEE International Conference on Automatic Face \&
  Gesture Recognition (2008)

\bibitem{Almeidaab14}
Almeidaab, S.G.M., Guimarãesc, F.G., Ramírez, J.: Feature extraction in
  brazilian sign language recognition based on phonological structure and using
  rgb-d sensors.
\newblock Expert Systems with Applications \textbf{41}(16), 7259--7271 (2014)

\bibitem{Athitsos08}
Athitsos, V., Neidle, C., Sclaroff, S., Nash, J., Stefan, A., Yuan, Q.,
  Thangali, A.: The asl lexicon video dataset.
\newblock In: Proceedings of CVPR 2008 Workshop on Human Communicative
  Behaviour Analysis. IEEE (2008)

\bibitem{buehler2011upper}
Buehler, P., Everingham, M., Huttenlocher, D.P., Zisserman, A.: Upper body
  detection and tracking in extended signing sequences.
\newblock International journal of computer vision \textbf{95}(2), 180 (2011)

\bibitem{camgoz2017subunets}
Camg{\"o}z, N.C., Hadfield, S., Koller, O., Bowden, R.: Subunets: End-to-end
  hand shape and continuous sign language recognition.
\newblock In: ICCV, vol.~1 (2017)

\bibitem{camgoz2018neural}
Camgoz, N.C., Hadfield, S., Koller, O., Ney, H., Bowden, R.: Neural sign
  language translation.
\newblock CVPR 2018 Proceedings  (2018)

\bibitem{I3D}
Carreira, J., Zisserman, A.: Quo vadis, action recognition? a new model and the
  kinetics dataset.
\newblock In: Computer Vision and Pattern Recognition (CVPR), 2017 IEEE
  Conference on, pp. 4724--4733. IEEE (2017)

\bibitem{Chai13}
Chai, X., Li, G., Lin, Y., Xu, Z., Tang, Y., Chen, X., Zhou, M.: Sign language
  recognition and translation with kinect.
\newblock In: Proceedings of IEEE International Conference on Automatic Face
  and Gesture Recognition (2013)

\bibitem{charles2014automatic}
Charles, J., Pfister, T., Everingham, M., Zisserman, A.: Automatic and
  efficient human pose estimation for sign language videos.
\newblock International Journal of Computer Vision \textbf{110}(1), 70--90
  (2014)

\bibitem{cui2017recurrent}
Cui, R., Liu, H., Zhang, C.: Recurrent convolutional neural networks for
  continuous sign language recognition by staged optimization.
\newblock In: IEEE Conference on Computer Vision and Pattern Recognition (CVPR)
  (2017)

\bibitem{T3D}
{Diba}, A., {Fayyaz}, M., {Sharma}, V., {Karami}, A.H., {Mahdi Arzani}, M.,
  {Yousefzadeh}, R., {Van Gool}, L.: {Temporal 3D ConvNets: New Architecture
  and Transfer Learning for Video Classification}.
\newblock ArXiv e-prints  (2017)

\bibitem{donahue2014long}
Donahue, J., Anne~Hendricks, L., Guadarrama, S., Rohrbach, M., Venugopalan, S.,
  Saenko, K., Darrell, T.: Long-term recurrent convolutional networks for
  visual recognition and description.
\newblock In: Proceedings of the IEEE conference on Computer Vision and Pattern
  Recognition, pp. 2625--2634 (2015)

\bibitem{donahue2013decaf}
Donahue, J., Jia, Y., Vinyals, O., Hoffman, J., Zhang, N., Tzeng, E., Darrell,
  T.: Decaf: A deep convolutional activation feature for generic visual
  recognition.
\newblock arXiv preprint arXiv:1310.1531  (2013)

\bibitem{DreuwECCV2010}
Dreuw, P., Forster, J., Ney, H.: Tracking benchmark databases for video-based
  sign language recognition.
\newblock In: Proc. ECCV International Workshop on Sign, Gesture, and Activity
  (2010)

\bibitem{Er-Rady17}
Er-Rady, A., Thami, R.O.H., Faizi, R., Housni, H.: Automatic sign language
  recognition: A survey.
\newblock In: Proceedings of the 3rd International Conference on Advanced
  Technologies for Signal and Image Processing (2017)

\bibitem{Fang07}
Fang, G., Gao, W., Zhao, D.: Large-vocabulary continuoius sign language
  recognition based on transition-movement models.
\newblock IEEE Transactions on Systems, Man, and Cybernetics - Part A: Systems
  and Humans \textbf{37}(1) (2007)

\bibitem{fernando2017rank}
Fernando, B., Gavves, E., Oramas, J., Ghodrati, A., Tuytelaars, T.: Rank
  pooling for action recognition.
\newblock IEEE transactions on Pattern Analysis and Machine Intelligence
  \textbf{39}(4), 773--787 (2017)

\bibitem{forster2012rwth}
Forster, J., Schmidt, C., Hoyoux, T., Koller, O., Zelle, U., Piater, J.H., Ney,
  H.: Rwth-phoenix-weather: A large vocabulary sign language recognition and
  translation corpus.
\newblock In: LREC, pp. 3785--3789 (2012)

\bibitem{Furman10}
Furman, N., Goldberg, D., Lusin, N.: Enrollments in languages other than
  english in united states institutions of higher education, fall 2010.
\newblock Retrieved from http://www.mla.org/2009\_enrollmentsurvey  (2010)

\bibitem{gattupalli2016evaluation}
Gattupalli, S., Ghaderi, A., Athitsos, V.: Evaluation of deep learning based
  pose estimation for sign language recognition.
\newblock In: Proceedings of the 9th ACM International Conference on Pervasive
  Technologies Related to Assistive Environments, p.~12. ACM (2016)

\bibitem{girshick2014rich}
Girshick, R., Donahue, J., Darrell, T., Malik, J.: Rich feature hierarchies for
  accurate object detection and semantic segmentation.
\newblock In: Computer Vision and Pattern Recognition (CVPR), 2014 IEEE
  Conference on, pp. 580--587. IEEE (2014)

\bibitem{GuyonCGD2011}
Guyon, I., Athitsos, V., Jangyodsuk, P., Escalante, H.J.: The chalearn gesture
  dataset (cgd 2011).
\newblock Machine Vision and Applications \textbf{25}(8), 1929--1951 (2014)

\bibitem{3DResNet}
Hara, K., Kataoka, H., Satoh, Y.: Can spatiotemporal 3d cnns retrace the
  history of 2d cnns and imagenet?
\newblock In: Proceedings of the IEEE Conference on Computer Vision and Pattern
  Recognition (CVPR), pp. 6546--6555 (2018)

\bibitem{he2014spatial}
He, K., Zhang, X., Ren, S., Sun, J.: Spatial pyramid pooling in deep
  convolutional networks for visual recognition.
\newblock In: Computer Vision--ECCV 2014, pp. 346--361. Springer (2014)

\bibitem{huang2018video}
Huang, J., Zhou, W., Zhang, Q., Li, H., Li, W.: Video-based sign language
  recognition without temporal segmentation.
\newblock arXiv preprint arXiv:1801.10111  (2018)

\bibitem{huenerfauth2017evaluation}
Huenerfauth, M., Gale, E., Penly, B., Pillutla, S., Willard, M., Hariharan, D.:
  Evaluation of language feedback methods for student videos of american sign
  language.
\newblock ACM Transactions on Accessible Computing (TACCESS) \textbf{10}(1), 2
  (2017)

\bibitem{3D}
Ji, S., Xu, W., Yang, M., Yu, K.: 3d convolutional neural networks for human
  action recognition.
\newblock IEEE transactions on pattern analysis and machine intelligence
  \textbf{35}(1), 221--231 (2013)

\bibitem{Jiang14}
Jiang, Y., Tao, J., Ye, W., Wang, W., Ye, Z.: An isolated sign language
  recognition system using rgb-d sensor with sparse coding.
\newblock In: Proceedings of IEEE 17th International Conference on
  Computational Science and Engineering (2014)

\bibitem{VideoYOLO}
Jing, L., Yang, X., Tian, Y.: Video you only look once: Overall temporal
  convolutions for action recognition.
\newblock Journal of Visual Communication and Image Representation \textbf{52},
  58--65 (2018)

\bibitem{3DMultiModel}
Jing, L., Ye, Y., Yang, X., Tian, Y.: 3d convolutional neural network with
  multi-model framework for action recognition.
\newblock In: 2017 IEEE International Conference on Image Processing (ICIP),
  pp. 1837--1841. IEEE (2017)

\bibitem{Kadous96}
Kadous, M.: Machine recognition of auslan signs using powergloves:towards
  large-lexicon recognition of sign language.
\newblock In: Proceedings of the Workshop on the Integration of Gesture in
  Language and Speech, pp. 165--174 (1996)

\bibitem{karpathy2014deep}
Karpathy, A., Fei-Fei, L.: Deep visual-semantic alignments for generating image
  descriptions.
\newblock arXiv preprint arXiv:1412.2306  (2014)

\bibitem{KarpathyCVPR14}
Karpathy, A., Toderici, G., Shetty, S., Leung, T., Sukthankar, R., Fei-Fei, L.:
  Large-scale video classification with convolutional neural networks.
\newblock In: CVPR (2014)

\bibitem{Kinetics}
Kay, W., Carreira, J., Simonyan, K., Zhang, B., Hillier, C., Vijayanarasimhan,
  S., Viola, F., Green, T., Back, T., Natsev, P., et~al.: The kinetics human
  action video dataset.
\newblock arXiv preprint arXiv:1705.06950  (2017)

\bibitem{DanielKelly10}
Kelly, D., McDonald, J., Markham, C.: A person independent system for
  recognition of hand postures used in sign language.
\newblock Pattern Recognition Letters \textbf{31}(11), 1359--1368 (2010)

\bibitem{Keskin12}
Keskin, C., Kıraç, F., Kara, Y., Akarun, L.: Hand pose estimation and hand
  shape classification using multi-layered randomized decision forests.
\newblock In: In Proceedings of the European Conference on Computer Vision, pp.
  852--863 (2012)

\bibitem{koller2015continuous}
Koller, O., Forster, J., Ney, H.: Continuous sign language recognition: Towards
  large vocabulary statistical recognition systems handling multiple signers.
\newblock Computer Vision and Image Understanding \textbf{141}, 108--125 (2015)

\bibitem{koller2015deep}
Koller, O., Ney, H., Bowden, R.: Deep learning of mouth shapes for sign
  language.
\newblock In: Proceedings of the IEEE International Conference on Computer
  Vision Workshops, pp. 85--91 (2015)

\bibitem{koller2016deep}
Koller, O., Ney, H., Bowden, R.: Deep hand: How to train a cnn on 1 million
  hand images when your data is continuous and weakly labelled.
\newblock In: Proceedings of the IEEE Conference on Computer Vision and Pattern
  Recognition, pp. 3793--3802 (2016)

\bibitem{koller2018deep}
Koller, O., Zargaran, S., Ney, H., Bowden, R.: Deep sign: Enabling robust
  statistical continuous sign language recognition via hybrid cnn-hmms.
\newblock International Journal of Computer Vision \textbf{126}(12), 1311--1325
  (2018)

\bibitem{Kong14}
Kong, W., Ranganath, S.: Towards subject independent continues sign language
  recognition: A segment and merge approach.
\newblock Pattern Recognition \textbf{47}(3), 1294--1308 (2014)

\bibitem{AlexNet}
Krizhevsky, A., Sutskever, I., Hinton, G.E.: Imagenet classification with deep
  convolutional neural networks.
\newblock In: Advances in Neural Information Processing Systems, pp. 1097--1105
  (2012)

\bibitem{Kumar17}
Kumar, P., Gauba, H., Roy, P.P., Dogra, D.P.: A multimodal framework for sensor
  based sign language recognition.
\newblock Neurocomputing \textbf{259}, 21--38 (2017)

\bibitem{kumar2018independent}
Kumar, P., Roy, P.P., Dogra, D.P.: Independent bayesian classifier combination
  based sign language recognition using facial expression.
\newblock Information Sciences \textbf{428}, 30--48 (2018)

\bibitem{Lang12}
Lang, S., Block, M., Rojas, R.: Sign language recognition using kinect.
\newblock In: In Proceedings of International Conference on Artificial
  Intelligence and Soft Computing, pp. 394--402 (2012)

\bibitem{Liang98}
Liang, R.H., Ouhyoung, M.: A real-time continuous gesture recognition system
  for sign language.
\newblock In: Proceedings of the Third IEEE International Conference on
  Automatic Face and Gesture Recognition, pp. 558--567 (1998)

\bibitem{JingjingLiu13}
Liu, J., Liu, B., Zhang, S., Yang, F., Yang, P., Metaxas, D.N., Neidle, C.:
  Recognizing eyebrow and periodic head gestures using crfs for non-manual
  grammatical marker detection in asl.
\newblock In: Proc. of the 10th IEEE International Conference and Workshops on
  Automatic Face and Gesture Recognition (FG) (2013)

\bibitem{liu2015rgbd}
Liu, W., Fan, Y., Li, Z., Zhang, Z.: Rgbd video based human hand trajectory
  tracking and gesture recognition system.
\newblock Mathematical Problems in Engineering \textbf{2015} (2015)

\bibitem{liu2016real}
Liu, Z., Huang, F., Tang, G.W.L., Sze, F.Y.B., Qin, J., Wang, X., Xu, Q.:
  Real-time sign language recognition with guided deep convolutional neural
  networks.
\newblock In: Proceedings of the 2016 Symposium on Spatial User Interaction,
  pp. 187--187. ACM (2016)

\bibitem{lu2012cuny}
Lu, P., Huenerfauth, M.: Cuny american sign language motion-capture corpus:
  first release.
\newblock In: Proceedings of the 5th Workshop on the Representation and
  Processing of Sign Languages: Interactions between Corpus and Lexicon, The
  8th International Conference on Language Resources and Evaluation (LREC
  2012), Istanbul, Turkey (2012)

\bibitem{RVL-SLLLASLDatabase02}
Martínez, A.M., Wilbur, R.B., Shay, R., Kak, A.C.: The rvl-slll asl database.
\newblock In: Proc. of IEEE International Conference Multimodal Interfaces
  (2002)

\bibitem{Mehrotra15}
Mehrotra, K., Godbole, A., Belhe, S.: Indian sign language recognition using
  kinect sensor.
\newblock In: In Proceedings of the International Conference Image Analysis and
  Recognition, pp. 528--535 (2015)

\bibitem{Metaxas12}
Metaxas, D., Liu, B., Yang, F., Yang, P., Michael, N., Neidle, C.: Recognition
  of nonmanual markers in asl using non-parametric adaptive 2d-3d face
  tracking.
\newblock In: Proc. of the Int. Conf. on Language Resources and Evaluation
  (LREC), European Language Resources Association (2012)

\bibitem{miao2017multimodal}
Miao, Q., Li, Y., Ouyang, W., Ma, Z., Xu, X., Shi, W., Cao, X., Liu, Z., Chai,
  X., Liu, Z., et~al.: Multimodal gesture recognition based on the resc3d
  network.
\newblock In: ICCV Workshops, pp. 3047--3055 (2017)

\bibitem{mitchell2006many}
Mitchell, R.E., Young, T.A., Bachleda, B., Karchmer, M.A.: How many people use
  asl in the united states? why estimates need updating.
\newblock Sign Language Studies \textbf{6}(3), 306--335 (2006)

\bibitem{Mulrooney10}
Mulrooney, K.: American Sign Language Demystified, Hard Stuff Made Easy.
\newblock McGraw Hill (2010)

\bibitem{narayana2018gesture}
Narayana, P., Beveridge, J.R., Draper, B.A.: Gesture recognition: Focus on the
  hands.
\newblock In: Proceedings of the IEEE Conference on Computer Vision and Pattern
  Recognition, pp. 5235--5244 (2018)

\bibitem{Neidle12}
Neidle, C., Thangali, A., Sclaroff, S.: Challenges in development of the
  american sign language lexicon video dataset (asllvd) corpus.
\newblock In: Proceedings of the Language Resources and Evaluation Conference
  (LREC) (2012)

\bibitem{neidle2012new}
Neidle, C., Vogler, C.: A new web interface to facilitate access to corpora:
  Development of the asllrp data access interface (dai).
\newblock In: Proc. 5th Workshop on the Representation and Processing of Sign
  Languages: Interactions between Corpus and Lexicon, LREC (2012)

\bibitem{Ong05}
Ong S. C.and~Ranganath, S.: Automatic sign language analysis: A survey and the
  future beyond lexical meaning.
\newblock IEEE Pattern Analysis and Machine Intelligence \textbf{27}(6),
  873--891 (2005)

\bibitem{palmeri2017sign}
Palmeri, M., Vella, F., Infantino, I., Gaglio, S.: Sign languages recognition
  based on neural network architecture.
\newblock In: International Conference on Intelligent Interactive Multimedia
  Systems and Services, pp. 109--118. Springer (2017)

\bibitem{Pigou14}
Pigou, L., Dieleman, S., Kindermans, P.J., Schrauwen, B.: Sign language
  recognition using convolutional neural networks.
\newblock In: Proceedings of European Conference on Computer Vision Workshops,
  pp. 572--578 (2014)

\bibitem{pigou2018beyond}
Pigou, L., Van Den~Oord, A., Dieleman, S., Van~Herreweghe, M., Dambre, J.:
  Beyond temporal pooling: Recurrence and temporal convolutions for gesture
  recognition in video.
\newblock International Journal of Computer Vision \textbf{126}(2-4), 430--439
  (2018)

\bibitem{pigou2017gesture}
Pigou, L., Van~Herreweghe, M., Dambre, J.: Gesture and sign language
  recognition with temporal residual networks.
\newblock In: Proceedings of the IEEE Conference on Computer Vision and Pattern
  Recognition, pp. 3086--3093 (2017)

\bibitem{pu2018dilated}
Pu, J., Zhou, W., Li, H.: Dilated convolutional network with iterative
  optimization for continuous sign language recognition.
\newblock In: IJCAI, pp. 885--891 (2018)

\bibitem{Pugeault11}
Pugeault, N., Bowden, R.: Spelling it out: Real-time asl fingerspelling
  recognition.
\newblock In: Proc. of IEEE International Conference on Computer Vision
  Workshops, pp. 1114--1119 (2011)

\bibitem{P3D}
Qiu, Z., Yao, T., Mei, T.: Learning spatio-temporal representation with
  pseudo-3d residual networks.
\newblock In: The IEEE International Conference on Computer Vision (ICCV)
  (2017)

\bibitem{RenYuan13}
Ren, Z., Yuan, J., Meng, J., Zhang, Z.: Robust part-based hand gesture
  recognition using kinect sensor.
\newblock IEEE Trans. on Multimedia \textbf{15}, 1110--1120 (2013)

\bibitem{simonyan2014two}
Simonyan, K., Zisserman, A.: Two-stream convolutional networks for action
  recognition in videos.
\newblock In: Advances in Neural Information Processing Systems, pp. 568--576
  (2014)

\bibitem{VGG}
Simonyan, K., Zisserman, A.: Very deep convolutional networks for large-scale
  image recognition.
\newblock arXiv preprint arXiv:1409.1556  (2014)

\bibitem{Starner98}
Starner, T., Weaver, J., Pentland, A.: Real-time american sign language
  recognition using desk and wearable computer based video.
\newblock IEEE Pattern Analysis and Machine Intelligence \textbf{20}(12),
  1371--1375 (1998)

\bibitem{szegedy2014going}
Szegedy, C., Liu, W., Jia, Y., Sermanet, P., Reed, S., Anguelov, D., Erhan, D.,
  Vanhoucke, V., Rabinovich, A.: Going deeper with convolutions.
\newblock arXiv preprint arXiv:1409.4842  (2014)

\bibitem{Tamura88}
Tamura, S., Kawasaki, S.: Recognition of sign language motion images.
\newblock Pattern Recognition \textbf{21}(4), 343--353 (1988)

\bibitem{C3D}
Tran, D., Bourdev, L., Fergus, R., Torresani, L., Paluri, M.: Learning
  spatiotemporal features with 3d convolutional networks.
\newblock In: Proceedings of the IEEE International Conference on Computer
  Vision, pp. 4489--4497 (2015)

\bibitem{traxler2000stanford}
Traxler, C.B.: The stanford achievement test: National norming and performance
  standards for deaf and hard-of-hearing students.
\newblock Journal of deaf studies and deaf education \textbf{5}(4), 337--348
  (2000)

\bibitem{Valli11}
Valli, C., Lucas, C., Mulrooney, K.J., Villanueva, M.: Linguistics of American
  Sign Language: An Introduction.
\newblock Gallaudet University Press (2011)

\bibitem{LargeScaleRGBDHandGesture16}
Wan, J., Li, S., Zhao, Y., Zhou, S., Guyon, I., Escalera, S.: Chalearn looking
  at people rgb-d isolated and continuous datasets for gesture recognition.
\newblock In: Proceedings of CVPR 2008 Workshops. IEEE (2016)

\bibitem{wang2017large}
Wang, H., Wang, P., Song, Z., Li, W.: Large-scale multimodal gesture
  recognition using heterogeneous networks.
\newblock In: Proceedings of the IEEE Conference on Computer Vision and Pattern
  Recognition, pp. 3129--3137 (2017)

\bibitem{YangSpotting09}
Yang, H., Sclaroff, S., Lee, S.: Sign language spotting with a threshold model
  based on conditional random fields.
\newblock IEEE Pattern Analysis and Machine Intelligence \textbf{31}(7),
  1264--1277 (2009)

\bibitem{YangSensors15}
Yang, H.D.: Sign language recognition with the kinect sensor based on
  conditional random fields.
\newblock Sensors \textbf{15}, 135--147 (2015)

\bibitem{YangSarkar10}
Yang, R., Sarkar, S., Loeding, B.: Handling movement epenthesis and hand
  segmentation ambiguities in continuous sign language recognition using nested
  dynamic programming.
\newblock IEEE Pattern Analysis and Machine Intelligence \textbf{32}(3),
  462--477 (2010)

\bibitem{Yuancheng18}
Ye, Y., Tian, Y., Huenerfauth, M.: Recognizing american sign language gestures
  from within continuous videos.
\newblock The 8th IEEE Workshop on Analysis and Modeling of Faces and Gestures
  (AMFG) in conjunction with CVPR 2018  (2017)

\bibitem{ng2015beyond}
Yue-Hei~Ng, J., Hausknecht, M., Vijayanarasimhan, S., Vinyals, O., Monga, R.,
  Toderici, G.: Beyond short snippets: Deep networks for video classification.
\newblock In: Proceedings of the IEEE conference on Computer Vision and Pattern
  Recognition, pp. 4694--4702 (2015)

\bibitem{Zafrulla11}
Zafrulla, Z., Brashear, H., Starner, T., Hamilton H.and~Presti, P.: American
  sign language recognition with the kinect.
\newblock In: In Proceedings of the International Conference on Multimodal
  Interfaces, pp. 279--286 (2011)

\bibitem{ChenyangZhang16}
Zhang, C., Tian, Y., Huenerfauth, M.: Multi-modality american sign language
  recognition.
\newblock In: Proceedings of IEEE International Conference on Image Processing
  (ICIP) (2016)

\bibitem{zhang2017learning}
Zhang, L., Zhu, G., Shen, P., Song, J., Shah, S.A., Bennamoun, M.: Learning
  spatiotemporal features using 3dcnn and convolutional lstm for gesture
  recognition.
\newblock In: Proceedings of the IEEE Conference on Computer Vision and Pattern
  Recognition, pp. 3120--3128 (2017)

\end{thebibliography}

\end{document}